\documentclass[acmsmall]{acmart} 

\usepackage{amsmath,amsfonts}
\usepackage{algorithmic}
\usepackage{array}
\usepackage{booktabs}
\usepackage{textcomp}
\usepackage{stfloats}
\usepackage{url}
\usepackage{verbatim}
\usepackage{graphicx}
\usepackage[caption=false,font=normalsize,labelfont=sf,textfont=sf]{subfig}

\usepackage{threeparttable}
\usepackage{listings}
\usepackage{graphics}
\usepackage{colortbl}
\usepackage{multirow}
\usepackage{multicol}
\usepackage{bbding}
\usepackage{microtype}
\usepackage{hyperref}
\usepackage{color}

\AtBeginDocument{%
  }

\setcopyright{acmlicensed}
\copyrightyear{2026}
\acmYear{2026}
\acmDOI{10.1145/3774887}

\acmJournal{TOMM}
\acmVolume{22}
\acmNumber{1}
\acmArticle{26}
\acmMonth{1} 

\begin{document}

\title{3DMambaComplete: Structured State Space Model for High-Efficiency Point Cloud Completion}


\author{Yixuan Li}
\email{yxli24@m.fudan.edu.cn}
\orcid{0000-0002-9229-7555}
\affiliation{%
  \institution{Shanghai Key Laboratory of Data Science, School of Computer Science, Fudan University}
  \city{Shang hai}
  \country{China}
}

\author{Lipeng Ma}
\email{lpma21@m.fudan.edu.cn}
\orcid{0000-0001-5974-5988}
\affiliation{%
  \institution{Shanghai Key Laboratory of Data Science, School of Computer Science, Fudan University}
  \city{Shang hai}
  \country{China}
}

\author{Weidong Yang}

\authornote{Corresponding author}
\email{wdyang@fudan.edu.cn}
\orcid{0000-0002-6473-9272}
\affiliation{%
  \institution{Shanghai Key Laboratory of Data Science, School of Computer Science, Fudan University}
  \city{Shang hai}
  \country{China}
}

\author{Ben Fei}
\authornotemark[1]
\email{benfei@cuhk.edu.hk}
\orcid{0000-0002-3219-9996}
\affiliation{%
  \institution{Department of Information Engineering, The Chinese University of Hong Kong}
  \city{Hong Kong}
  \country{China}
}

\renewcommand{\shortauthors}{Li et al.}

\begin{abstract}

Point cloud completion seeks to reconstruct a complete and high-fidelity point cloud from an incomplete and low-quality input. 
Current methods predominantly rely on Transformer architectures for feature extraction.
However, these approaches face two major limitations, including the computational complexity associated with the attention mechanism and the potential loss of fine-grained details during pooling operations. These issues hinder their performance on large-scale and highly fragmented point clouds.
To overcome these challenges, we propose 3DMambaComplete, a novel point cloud completion method based on the selective State Space Model (SSM), particularly leveraging the Mamba architecture.
Unlike traditional Transformer-based methods, 3DMambaComplete utilizes Mamba's linear-time complexity to efficiently extract global features with significantly reduced computational overhead.
Furthermore, we introduce the concepts of discriminative nodes, referred to as hyperpoints, along with dynamic offsets, to improve reconstruction quality.
Specifically, the HyperPoint Generation Module encodes the downsampled features of the point cloud using the Mamba Encoder, producing a set of hyperpoints that capture critical information. 
Subsequently, the HyperPoint Spread Module disperses these hyperpoints across various spatial locations employing dynamic offsets to mitigate aggregation. 
Finally, the Point Deformation Module implements a deformation technique to transform the 2D mesh into a detailed 3D structure, resulting in high-quality point cloud completions.
Experiments on widely-used benchmark datasets show that 3DMambaComplete outperforms existing point cloud completion techniques in both quantitative and qualitative evaluations.
\end{abstract}




\maketitle

\section{Introduction}

Point clouds are a fundamental representation of three-dimensional geometries acquired through 3D sensors such as laser scanners, LiDAR, and depth cameras. They provide a comprehensive description of the distinct properties exhibited by real-world objects, making them invaluable in diverse fields such as robotics, computer graphics, and industrial inspection~\cite{wang2025compression,fei2022comprehensive,zhang2025mmpcqa}.
However, capturing point clouds in real-world scenarios often poses challenges, including noise, occlusion, sensor limitations, and reconstruction errors. Consequently, the resulting point clouds are often incomplete and distorted~\cite{fei2023dctr,mittal2022autosdf,wang2022learning}.
To facilitate subsequent tasks effectively, a crucial step involves the completion or reconstruction of incomplete point clouds to restore their information and enhance the quality of objects~\cite{zhang2024learning}.
Nonetheless, the sparse and unstructured nature of point clouds, coupled with the absence of a fixed spatial arrangement, presents substantial challenges for reconstruction efforts~\cite{mittal2022autosdf,wang2022learning,yan2022shapeformer}.

Previous research has applied deep learning methods for point cloud downstream tasks, such as 3D CNN-based methods~\cite{rudolph2024transcoding,zhou2024multi,xie2020grnet}. However, these approaches are often constrained by high resolution and substantial computational demands. 
As a result, Point-based approaches, pioneered by PointNet~\cite{qi2017pointnet} and PointNet++~\cite{qi2017pointnet++}, have emerged as an alternate strategy, fostering the development of additional point-based completion methods~\cite{huang2020pf, xia2021asfm, wang2020cascaded}.
Drawing inspiration from the achievements of Transformers, recent approaches~\cite{xiang2021snowflakenet,yu2021pointr,chen2023anchorformer,li2023proxyformer,zhou2022seedformer} have redefined point cloud reconstruction as a mapping task between partially observed input spaces and their corresponding fully reconstructed output spaces. 
These approaches utilize encoder-decoder architectures with Transformer-based backbones to effectively reconstruct point clouds. However, Transformer-based methods face two critical challenges: (i) the quadratic complexity of attention mechanisms, which limits scalability to extended sequences, and (ii) the potential loss of local information due to pooling operations during encoding~\cite{wu2023mpct,cao2022vdtr}.

Recently, the State Space Model (SSM)~\cite{gu2021efficiently}, along with the Structured SSM and Mamba~\cite{gu2023mamba} approaches, has come up with a promising solution to address the inherent challenges faced by Transformers.
Mamba~\cite{gu2023mamba}, an efficient hardware-aware algorithm based on Structured SSM with time-varying parameters, demonstrates exceptional performance in handling long-range and discrete data, particularly in 2D vision applications such as segmentation and classification~\cite{gu2023mamba}.
As a pioneering work, PointMamba~\cite{liang2024pointmamba} extends the Mamba framework to encompass point cloud processing, aiming to overcome Mamba's limitation regarding the causality of input data and successfully capture point cloud information.
Motivated by this advancement, this paper proposes 3DMambaComplete, which applies the Mamba framework to the domain of point cloud completion. 
3DMambaComplete involves encoding and enhancing the downsampled points, generating a set of hyperpoints, learning specific offsets, and dispersing the generated hyperpoints to different 3D locations. 
Finally, by employing a deformation scheme, the new hyperpoints, combined with the downsampled points, undergo a transformation from a 2D representation to cohesive 3D structures, serving the purpose of point cloud completion.

In summary, 3DMambaComplete makes three main contributions:
\begin{itemize}
    \item We present 3DMambaComplete, a point cloud completion network built upon the 3DMamba architecture. 
    This method effectively completes large sequences by combining linear complexity and a global receptive field. 
    The selectivity of Mamba improves contextual comprehension and maintains vital local information for precise reconstruction.  
    \item We propose a HyperPoint Generation module to generate a new representation of shapes called Hyperpoints. 
    Subsequently, the Mamba Blocks enhance the features of sampled points and predict hyperpoints. 
    The HyperPoint Spread module disperses the hyperpoints, while the Point Deformation module transforms the points into high-quality 3D structures. 
    Additionally, a compactness constraint function is designed to guide the model in more effective learning.
    \item Extensive studies show that 3DMambaComplete enhances the quality of point cloud reconstruction, particularly for highly incomplete shapes, surpassing existing methods in evaluations conducted on PCN, KITTI, and ShapeNet34/55 datasets.
\end{itemize}

The overall organization of this paper is listed as follows:
Section 2 delineates related work on point cloud completion and state space models.
Section 3 introduces our novel 3DMambaComplete architecture.
Section 4 conducts extensive experiments across benchmark datasets, including ablation studies to comprehensively evaluate completion performance.
Section 5 concludes and outlines future research directions.

\section{Related Works}

\subsection{Point Cloud Completion}
Point cloud completion aims to recover complete 3D shapes from partial observations, a task that is particularly essential in real-world applications such as robotics, autonomous driving, and augmented reality. Traditional completion approaches relied on geometric priors or optimization-based algorithms~\cite{zhang2021view,tesema2023point}, but these methods often suffer from limited generalizability and high computational costs. In contrast, deep learning-based methods have demonstrated superior performance in handling noisy and incomplete data~\cite{fei2022comprehensive,lyu2021conditional,wang2022learning, spurek2022hyperpocket}. Early works adopted autoencoders~\cite{wang2020cascaded,tchapmi2019topnet} and generative models like GANs~\cite{achlioptas2018learning,zhang2021unsupervised} and diffusion models~\cite{luo2021diffusion,zhou20213d} to learn mappings from partial inputs to complete shapes.
For instance, ASFM-Net~\cite{xia2021asfm} completed imcomplete and full point clouds by mapping them onto a common latent space. 
Later on, studies using GANs~\cite{achlioptas2018learning,zhang2021unsupervised} or DDPM ~\cite{luo2021diffusion,zhou20213d} also showed encouraging results.
Zhang et al.~\cite{zhang2021unsupervised} used GAN inversion to complete 3D shapes, whereas Luo et al.~\cite{luo2021diffusion} generated point clouds by applying non-equilibrium thermodynamic diffusion processes.

With the growing success of Transformers in vision and NLP domains~\cite{han2022survey,li2022contextual,latif2023transformers}, researchers have extended their use to point cloud processing~\cite{yu2021pointr,pang2022masked, zhang2022qinet, zhang2024compoint}.
For example, SnowflakeNet~\cite{xiang2021snowflakenet} adopt hierarchical upsampling schemes to generate dense point clouds progressively. To better capture the geometric context, intermediate structures are introduced, including seeds~\cite{zhou2022seedformer}, anchors~\cite{chen2023anchorformer}, and proxies~\cite{li2023proxyformer}. 
Despite their effectiveness, these methods face scalability challenges and struggle to retain fine-grained details due to pooling operations and token compression~\cite{fei2022vq, fei2024progressive}. 
Recent works attempt to mitigate these issues. 
CRA-PCN~\cite{rong2024cra}  introduces a cross-resolution Transformer with local attention for intra- and inter-level feature aggregation, enhancing multi-scale detail capture.
In contrast, MMDR~\cite{fei2025multi} adopts a multimodal approach, which employs differentiable rendering to generate depth images as supervision signals for precise shape learning. The architecture integrates point-based rendering, attentive feature extraction, and PU-Transformer components, enforcing multimodal consistency between 3D and 2D cues for high-fidelity reconstruction. 
Nevertheless, the inherent quadratic complexity of the Transformer attention mechanism remains a bottleneck for efficiency and scalability. 
To address this, we propose 3DMambaComplete, a novel framework based on the SSM. 
By replacing self-attention with the Mamba architecture, our method achieves linear complexity in modeling long-range dependencies. 
Moreover, the selective state update mechanism enables simultaneous encoding of global context and fine-grained local geometry, mitigating information degradation caused by traditional pooling or downsampling operations.

\subsection{State Space Models}

State Space Models (SSMs)~\cite{heo2016open} have emerged as an effective framework for modeling long sequences, offering improved handling of long-range dependencies. For instance, HiPPO~\cite{gu2020hippo} combines deep learning with linear state space equations. Building on this, Gu et al. proposed the Linear State Space Learning (LSSL) method~\cite{gu2021combining}, which discretizes continuous-time SSMs and integrates linear state space equations with neural networks to enhance time series modeling. The S4 model~\cite{gu2021efficiently} introduces structural priors into sequence modeling and employs diagonalization of parameters to capture long-range dependencies. GSS~\cite{mehta2022long} simplifies SSMs through a gated activation mechanism, reducing model dimensionality and improving practicality. Further simplifying S4, Smith et al. introduced S5~\cite{smith2022simplified}, designed for greater usability in real-world scenarios.

More recently, Mamba~\cite{gu2023mamba} incorporates selection mechanisms and hardware-aware algorithms, enabling efficient linear-time inference and training across diverse domains. MoE-Mamba~\cite{pioro2024moe} combines Mamba with a Mixture of Experts (MoE) to boost expressiveness, state selection, and generalization. MambaByte~\cite{wang2024mambabyte} operates directly on byte sequences, avoiding tokenization overhead and improving accuracy and efficiency. GraphMamba~\cite{wang2024graph} extends sequential modeling to graph-structured data via node prioritization and ordering, improving contextual reasoning.
Notably, Mamba has achieved significant progress in biomedical imaging~\cite{ruan2024vm,liu2024swin,ye2024p,guo2024mambamorph}, where its ability to capture global context efficiently benefits complex visual tasks. It has also shown promise in 3D point cloud processing. For example, PointMamba~\cite{zhang2025point} uses spatial curve serialization to improve classification and segmentation, and Mamba3D~\cite{han2024mamba3d} employs local feature extraction and bidirectional SSMs to enhance representation learning.
Despite these advances, point cloud completion remains under-explored with Mamba. In this work, we adapt the Mamba architecture specifically for point cloud completion. To address the limitations of Transformers, such as quadratic complexity and limited local feature modeling. We leverage Mamba’s selective state space capabilities and global receptive field. We further propose a HyperPoint Spread Module and a Point Deformation Module to reconstruct high-quality complete point clouds.

\section{3DMambaComplete}
\begin{figure*}[t]
    \centering
    \includegraphics[width=\linewidth]{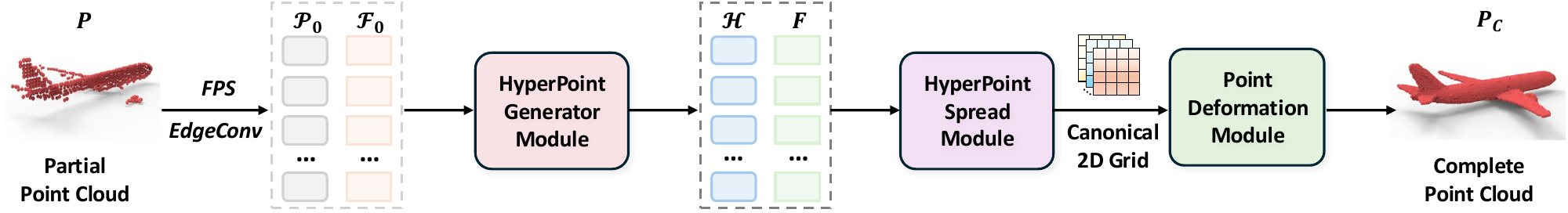}
    \caption{A brief overview of 3DMambaComplete.
    3DMambaComplete has three primary modules:
    (1) \textbf{HyperPoint Generation}: Given a partial point cloud $\mathbf{P}$, the Mamba Encoder and Cross-Attention enhance the downsampled point cloud features $\mathcal{F}_0$ and $\mathcal{P}_0$ to gain the coordinates of hyperpoint $\mathcal{H}$, as well as labeling its features $\mathcal{F}$.
    (2) \textbf{HyperPoint Spread Module}: This module leverages offsets to rationally disperse the generated hyperpoints to various locations in space.
    (3) \textbf{Point Deformation}: The Mamba Decoder serves in this module to transform the downsampled points and hyperpoints into a new set of hyperpoints. A deformation module then converts the newly formed hyperpoints from a 2D grid to a 3D structure.}
    \label{fig:overview}
\end{figure*}

In this section, we provide a concise overview of the fundamental theory underlying the state-space model (S4) and the S6 model, and examine how Mamba can effectively mitigate the computational complexity associated with transformer-based attention mechanisms. Subsequently, we describe the pipeline of 3DMambaComplete in detail, as illustrated in Figure~\ref{fig:overview}. Following this, we present an in-depth discussion of each component of 3DMambaComplete, including the HyperPoint Generation (Section \ref{Section 3.2}), the HyperPoint Spread Module (Section \ref{Section 3.3}), the Point Deformation Module (Section \ref{Section 3.4}) and Loss Function (Section \ref{Section 3.5}).

\subsection{Preliminaries}
\label{sec:3.1}

\subsubsection{State Space Model}
Inspired by specific continuous systems~\cite{chen2023anchorformer}, SSMs establish a sequence-to-sequence transformation $\mathit{x}(t)\rightarrow \mathit{y}(t)$ with a set of first-order differential equations~\cite{gu2021efficiently}. 
However, when applied to real-world scenarios, SSMs are mainly tailored for continuous sequences but have difficulties with discrete input data, such as text or point clouds. 
Consequently, before SSMs can be effectively utilized, it is necessary to discretize them. 
This discretization process can be accomplished by employing the following formulas:
\begin{equation}
\begin{aligned}
&\bar{\mathit{A}}=exp(\Delta\mathit{A}),\\
&\bar{\mathit{B}}=(\Delta\mathit{A})^{-1}(exp(\Delta\mathit{A})-\mathit{I})\cdot \Delta\mathit{B},\\
&\bar{\mathit{C}}=\mathit{C},
\end{aligned}
\end{equation}
where $\Delta$ is a step size, $\mathit{A}$ represents the current state, $\mathit{B}$ describes the impact of the input on state transitions, and $\mathit{C}$ captures the transformation of the state at a specific moment into an output.
When the parameters undergo a transformation, a valid recurrence relation incorporating the hidden states $\mathit{h}_l$ is devised through linear recursion.
The recurrence form can be expressed as follows:
\begin{equation}
\begin{aligned}
&{\mathit{h}}_l=\mathit{\bar{\mathit{A}}}\mathit{h}_{l-1}+\mathit{\bar{\mathit{B}}}\mathit{x}_l,\\
&\mathit{y}_l=\mathit{\mathit{C}}\mathit{h}_l,
\end{aligned}
\end{equation}
where the input $\mathit{x}_l$ represents a sample of the underlying continuous signal $\mathit{x}(t)$, and $\mathit{x}_l=x(l\Delta)$. And
$\bar{\mathit{A}}\in R^{N \times N}$, $\bar{\mathit{B}}\in R^{N \times 1}$.
Although S4 decreases the computational complexity by linear transformation and discretization, its fixed matrix is not adaptive to input variance, limiting its application in point cloud processing. Specifically, the information of each point of point cloud data is not equal, especially when dealing with incomplete point clouds, the importance of each token needs to be dynamically adjusted according to the local information in order to reconstruct the complete structure, but the fixed matrix design of S4 can not realize this requirement.

\subsubsection{Selective SSMs} 
Mamba~\cite{gu2023mamba}, also designated as S6, introduces a novel concept of selective State Space Model that addresses the issue of the invariant matrix $\Delta$, $\mathit{A}$, $\mathit{B}$, and $\mathit{C}$ in S4. 
This model excels in tasks involving context comprehension and effectively filters input sequences. Specifically, Mamba transforms the traditional time-invariant SSM into a time-varying model by introducing a length dimension into the matrix, significantly enhancing the selectivity of the SSM.

Building upon this, we propose a 3DMambaComplete, a point cloud completion technique based on Mamba. 
This technique successfully transfers Mamba's application of selective information processing in the field of NLP to the domain of point cloud completion. 
3DMambaComplete leverages the selectivity and hardware awareness capabilities of S6 to perform linear complexity feature extraction on incomplete point cloud inputs. 
3DMambaComplete not only reduces the computational complexity but also maintains a strong feature extraction capability, enabling effective modeling of long sequence point clouds.

\subsection{HyperPoint Generation}
\label{Section 3.2}

\begin{figure*}[t]
    \centering
    \includegraphics[width=0.95\linewidth]{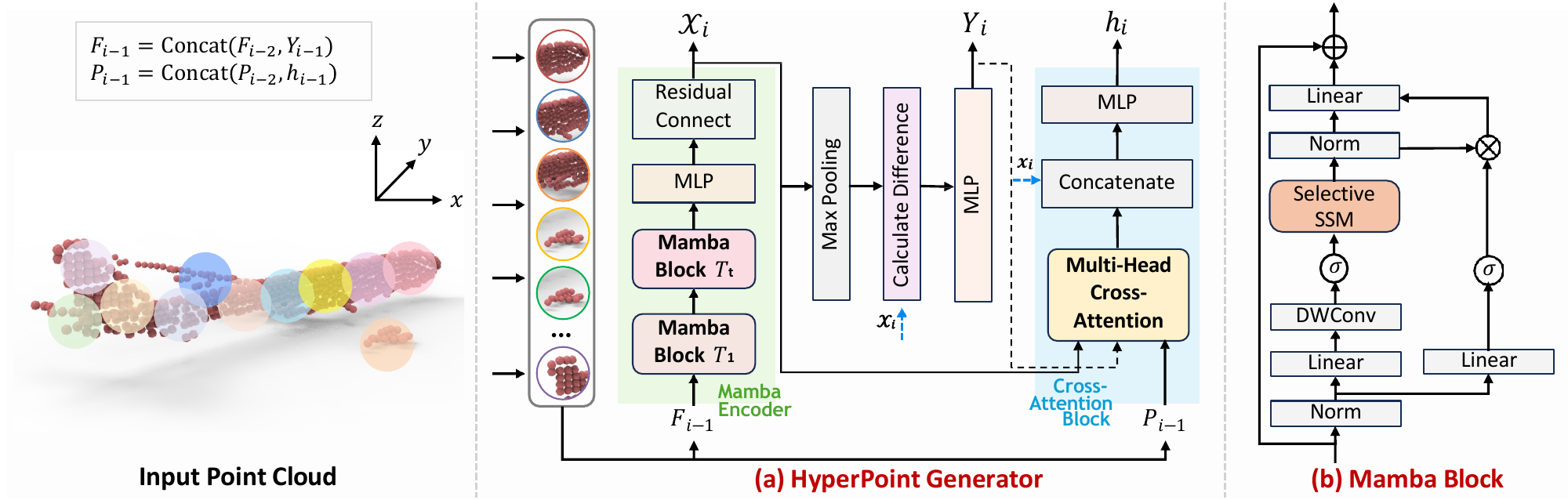}
    \caption{Illustration of the HyperPoint Generator Module. For the $i$-th Cross-Attention Block: (1) Input features $\mathcal{F}_{i-1}$ are enhanced by the Mamba Encoder (via $t$ Mamba Blocks) to produce $\mathcal{X}_i$; (2) Pooling derives $x_i$, with the disparity between $\mathcal{X}_i$ and $x_i$ extracting hyperpoint features $Y_i$; (3) These features interact with coordinates $\mathcal{P}_{i-1}$ through Cross-Attention to generate hyperpoint coordinates $h_i$.}
    \label{fig:hyperpoint_generation}
\end{figure*}

Transformer-based point cloud completion methods still face significant challenges when dealing with highly incomplete or large-scale point clouds. For highly incomplete inputs, the sparse number of points reduces computational cost, but the lack of sufficient geometric structure makes it difficult for the Transformer to infer missing regions. The fragmented nature of the input prevents the self-attention mechanism from capturing meaningful long-range dependencies, leading to poor completion results.
In large-scale scenarios, the quadratic complexity of self-attention with respect to the number of tokens introduces substantial training and inference overhead. Existing approaches often reduce the number of tokens through aggregation strategies, but this commonly results in the loss of fine-grained local geometry, degrading completion quality.

To address these challenges, we propose 3DMambaComplete, which introduces a Mamba-based hyperpoint generation module. By replacing self-attention with SSMs, our method significantly reduces computational complexity while maintaining the ability to capture global context. Additionally, we propose an intermediate representation named Hyperpoint that pools coordinate features from sampled points and combines them with generated hyperpoints using local modeling strategies. This design alleviates information loss commonly seen in traditional pooling operations, thereby improving geometric detail reconstruction during completion. 3DMambaComplete consists of three main modules, as shown in Figure~\ref{fig:overview}. 
Specifically, the process of generating hyperpoints involves two components: the Mamba Encoder and the Cross-Attention Block.

\textbf{Mamba Encoder.} 
Figure~\ref{fig:hyperpoint_generation} (a) illustrates the HyperPoint Generator Module.
Firstly, the incomplete point cloud input is downsampled by farthest point sampling (FPS)~\cite{qi2017pointnet}, resulting in $\mathit{N}$ sample points. 
These points are then subjected to feature extraction using EdgeConv:
\begin{equation}
\begin{aligned}
  &\mathcal{P}_0 = \mathit{FPS(P)},\\
  &\mathcal{F}_0 = \mathit{EdgeConv(\mathbf{\mathit{P}},\mathcal{P}_0)},
\end{aligned}
\end{equation}
where $\mathcal{P}_0 \in R^{N \times 3}$ denotes the downsampled points and the corresponding features serve as $\mathcal{F}_0 \in R^{N \times C}$.
Then, we design an encoder structure based on Mamba Block and Cross-Attention Block. The architecture of Mamba Block is shown in Figure~\ref{fig:hyperpoint_generation} (b), which contains layer normalization (LN), spatial smoothness modeling, depth-wise convolution (DW)~\cite{chollet2017xception}, and residual connections. 

For the $i$-th Cross-Attention Block, the input point features $\mathcal{F}_{i-1} \in \mathbb{R}^{N_{i-1} \times C}$ are converted into sequences by the $t$-th Mamba Block to obtain the processed features $\mathcal{Z}_{i -1}^{t}$ and reorder them according to a simple reordering strategy~\cite{liang2024pointmamba}. Afterward, these features are further processed by the MLP and residual connection to obtain the enhanced point cloud features $\mathcal{X}_{i} \in \mathbb{R}^{N_{i-1} \times C}$.
The following equations describe in detail the operation of the Mamba Block and its output:
\begin{align}
\mathcal{Z}_{i-1}^{'} &=\mathit{DW(MLP(LN(\mathcal{F}_{i-1})))},\\
\mathcal{Z}_{i-1}^{''} &=\mathit{MLP(LN(SSM(\sigma(\mathcal{Z}_{i-1}^{'}))))},\\
\mathcal{Z}_{i-1}^{t} &= \mathcal{Z}_{i-1}^{''} \times \sigma(LN(\mathcal{F}_{i-1}))+\mathcal{F}_{i-1},
\end{align}

The $\mathit{SSM}$ module plays a crucial role within the Mamba Blocks, and a comprehensive description of this module is provided in Section~\ref{sec:3.1}.
Later on, the enhanced point features are applied to the following generation of hyperpoints.

\textbf{Cross-Attention Block.} 
Inspired by~\cite{pilault2024block}, we employ a cross-attention mechanism to further process the enhanced downsampled point cloud features obtained from the Mamba Encoder, generating a set of new hyperpoints.
The utilization of the multi-head cross-attention mechanism allows improved handling of the enhanced point cloud feature values generated through the Mamba Encoder by dynamically processing the input point cloud sequence.

Figure~\ref{fig:hyperpoint_generation} provides a detailed depiction of the Cross-Attention Block within the hyperpoint generation process.
The enhanced features of the downsampled points are processed through max pooling and MLP layers, enabling the prediction of features for $\mathit{M}$ hyperpoints.
Following that the pooling feature vector $x_i$ is acquired, and its difference with $\mathcal{X}_{i}$ is computed. This difference is applied to enhance the downsampled point cloud features and the subsequent bias representation. The generated hyperpoint features
$Y_{i} \in \mathbb{R}^{M \times C}$ are learnt by:
\begin{align}
x_i &= \mathit{MaxPool}(\mathcal{X}_{i}), \\
Y_{i} &= \mathit{MLP}({x_i} - \mathcal{X}_{i}),
\end{align}
where the multi-layer perception and max pooling operation are denoted by $\mathit{MLP}(\cdot)$ and $\mathit{MaxPool}(\cdot)$, respectively.
The pooled feature vector $x_i$ is then added to the resultant combination in order to forecast the coordinates $h_{i} \in \mathbb{R}^{M \times 3}$ for the $M$ hyperpoints.
\begin{equation}
  h_i = \mathit{MaxPool(Concat[CrossAtten(
  {Y}_i,\mathcal{X}_{i},\mathcal{P}_{i-1}),\mathit{x}_i])},
\end{equation}
where $\mathit{Concat(\cdot)}$ and $\mathit{CrossAtten(\cdot)}$ serve as feature concatenation and the multi-head cross-attention. 
Furthermore, we utilize the cross-attention mechanism to compute the weights between 
$X_i$ and $Y_{i}$. The weights are then maintained 
for the purpose of combining the input points $\mathcal{F}_{i} \in \mathbb{R}^{N_i\times C}$ (where $N_i = N_{i-1} + M$).
Correspondingly, the input point $\mathcal{P}_{i-1}$ undergoes fusion with the predicted hyperpoints $h_i$, thereby forming the subsequent input point $\mathcal{P}_{i}$ as perceived by the $(i+1)$-th Cross-Attention Block.

\subsection{HyperPoint Spread Module}
\label{Section 3.3}

\begin{figure}[t]
    \centering
    \includegraphics[width=0.6\linewidth]{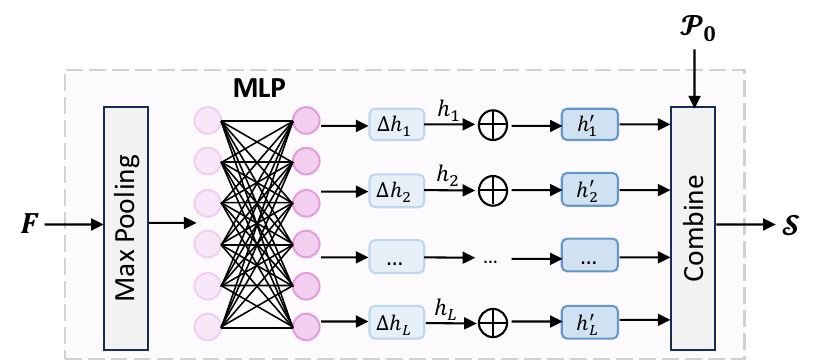}
    \caption{Detailed structure for HyperPoint Spread Module.}    
    \label{fig:hyperpoint_spread}
\end{figure}

Notably, we found that while the integration of Mamba Block and Cross-Attention Blocks enhances the extraction of point cloud features, the generated hyperpoints tend to concentrate in some specific regions. 
This concentration of hyperpoints may lead to relatively unsatisfactory geometries. 
To solve this problem, we need to utilize all the hyperpoints $\mathcal{H} \in \mathbb{R}^{L \times 3}$ obtained by continuous iterative enhancement in the Hyperpoints Generator Module through $i$ times Cross-Attention Block and multi-layer Mamba Block. 
After combining its features with the downsampled point features, we obtain $\mathit{F} \in \mathbb{R}^{(N+L) \times 3}$. 
The concept of offsets is introduced to effectively extend the hyperpoints to the entire space of the point cloud (including the unoccupied space), thus improving the accuracy of shape reconstruction.
This process is vividly illustrated in Figure~\ref{fig:hyperpoint_spread}.
We merge the hyperpoint features generated by the Cross-Attention Block with the improved downsampled point cloud features obtained from the Mamba Encoder to obtain the concatenated point features.
Subsequently, we predict the offsets for the hyperpoints and employ these offsets to assign each hyperpoint to a unique location in three-dimensional space. 
This approach effectively addresses the challenges associated with reconstruction that arise due to the limited number of hyperpoints.
The formulas can be expressed as follows:
\begin{align}
\mathit{F} &= \mathit{Con}(\mathbf{X}, \mathcal{F}_{H}), \\
\Delta\mathcal{H} &= \mathit{MLP}(\mathit{MaxPool}(\mathit{F})), \\
\mathcal{H}^{'} &= \mathcal{H} + \Delta\mathcal{H},
\end{align}

where $\mathbf{X}$ is the last Mamba Encoder enhanced feature under cross-attention, \(\mathcal{F}_{H} \in \mathbb{R}^{L \times 3}\) is all hyperpoint's features, \(\Delta\mathcal{H} \in \mathbb{R}^{M \times 3}\) is the offsets of hyperpoint, and \(\mathcal{H}^{'}\) is the spread hyperpoints.

\subsection{Point Deformation Module}

\label{Section 3.4}

Due to the previous focus on global features rather than considering the point-to-point relationships in point cloud completion methods~\cite{fei2022comprehensive,yang2018foldingnet}, the resulting deformation of a two-dimensional grid into a three-dimensional structure is often insufficiently detailed. 
Inspired by~\cite{xie2020grnet,chen2023anchorformer}, we integrate and leverage the correlation between local features of point clouds and hyperpoints for the purpose of 3D point cloud reconstruction.

\begin{figure}[t]
    \centering
    \includegraphics[width=0.6\linewidth]{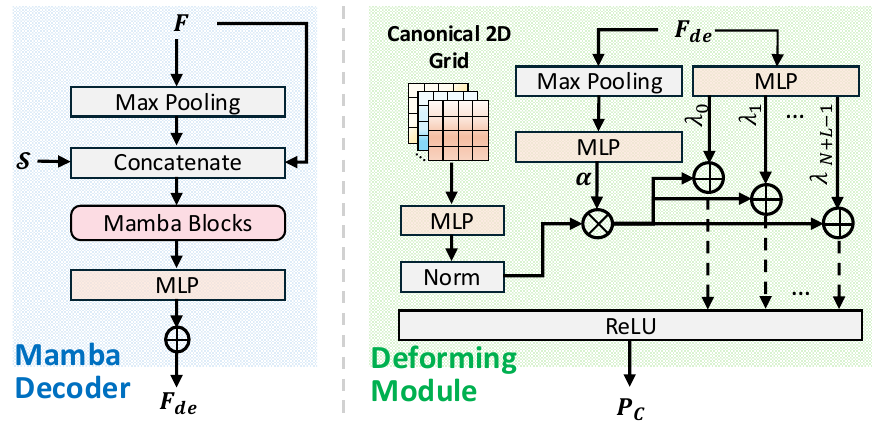}
    \caption{The pipeline of the Point Deformation Module.}
    \label{fig:point_deformation}
\end{figure}

Consequently, we propose a point deformation method to regulate the positional deformation of each hyperpoint.
Our Point Deformation module comprises two sub-modules: the Mamba Decoder and the deforming model, as illustrated in Figure~\ref{fig:point_deformation}.

The newly extracted hyperpoints $\mathcal{S}$ and their corresponding features are input into the Mamba Decoder, which consists of MLP layers and six Mamba Blocks.  
This results in the output feature $\mathit{F}_{de}$. 
And local point reconstruction is accomplished by feeding these characteristics into three successive point deformation modules.
Particularly, we compute the output feature vector of the new $\mathit{N+L}$ hyperpoints consisting of the downsampled points and the hyperpoints, along with the global point feature vector $\alpha$ and the local point feature vector $\lambda \in \mathbb{R}^{(N+L) \times C_l}$, where $\lambda = {\lambda}_{j=0}^{N+L-1}$. 
The formulas for these calculations are as follows:
\begin{equation}
    \alpha = \mathit{MLP(MaxPool(\mathit{F}_{de}))},  \lambda = \mathit{MLP(\mathit{F}_{de})},
\end{equation}
where the output dimension of $\mathit{MLP}$ in the $\mathit{l}$-th deforming block is $\mathit{C}_l$.
Both global features and local point features function as affinity factors in reconstructing the information for each new hyperpoint. 
When the input consists of 2D grid features $\mathit{g}_{in} \in \mathbb{R}^{K \times C_l}$, the output grid feature $\mathit{g}_{out} \in \mathbb{R}^{K \times C_l}$ can be computed using the following equation:
\begin{equation}
\mathit{g}_{out} = \lambda_j + \frac {\mathit{g}_{in} - \mu}{{\sigma}}\alpha,
\end{equation}
where $\mu$ and $\sigma$ represent the mean and standard deviation of input grid feature $\mathit{g}_{in}$. 

\subsection{Loss Function}
\label{Section 3.5}
The end-to-end 3DMambaComplete framework accomplishes the regularization of generated point clouds by optimizing the reconstruction loss and incorporating extended constraints.
Initially, the shapes generated by 3DMambaComplete undergo regularization using the Chamfer Distance (CD), effectively addressing the issue of unordered point clouds. 
Subsequently, 
the primary objective is to minimize the disparity between the generated point clouds and the ground truth, effectively reducing the potential for erroneous loss function formation.
Explicitly, $\mathcal{Q}$ designates the $n_{\mathcal{Q}}$ new hyperpoints that were predicted in 3DMambaComplete with hypepoints and sampled points, $\mathcal{P}$ signifies the $n_{\mathcal{P}}$ points of the predicted point cloud, and $\mathcal{G}$ symbolizes the ground truth, followed by the reconstructed point cloud loss $\mathcal{L}_{rec}$ may be computed as follows:
\begin{equation}
    \mathcal{L}_{CD_{qg}}={\dfrac{1}{n_{\boldsymbol{\mathcal{Q}}}}}{\sum_{q{\in}\boldsymbol{\mathcal{Q}}}}{\underset{g{\in}\boldsymbol{\mathcal{G}}}{min}}{\lVert}q-g{\rVert}+{\dfrac{1}{n_{\boldsymbol{\mathcal{G}}}}}{\sum_{g{\in}\boldsymbol{\mathcal{G}}}}{\underset{c{\in}\boldsymbol{\mathcal{Q}}}{min}}{\lVert}g-q{\rVert},
\end{equation}

\begin{equation}
    \mathcal{L}_{CD_{pg}}={\dfrac{1}{n_{\boldsymbol{\mathcal{P}}}}}{\sum_{p{\in}\boldsymbol{\mathcal{P}}}}{\underset{g{\in}\boldsymbol{\mathcal{G}}}{min}}{\lVert}p-g{\rVert}+{\dfrac{1}{n_{\boldsymbol{\mathcal{G}}}}}{\sum_{g{\in}\boldsymbol{\mathcal{G}}}}{\underset{p{\in}\boldsymbol{\mathcal{P}}}{min}}{\lVert}g-p{\rVert},
\end{equation}

\begin{equation}
\mathcal{L}_{rec}=\mathcal{L}_{CD_{cg}}+\mathcal{L}_{CD_{pg}},
\end{equation}

Furthermore, a minimum spanning tree is constructed on the new hyperpoints using a constrained loss function to ensure compactness of the points within the complemented space. 
The expansion loss, denoted as $\mathcal{L}_{expan}$, is derived from the previously mentioned minimum spanning tree loss, $\mathcal{L}_{tree}$. 
The mathematical representations of these functions are as follows:
\begin{equation}
\mathcal{L}_{tree}(\mathcal{P}_c,\zeta)={\sum_{(u,v){\in}{\mathcal{T}(\mathcal{P}_c)}}\mathbb{I}{\lbrace}d_{uv}\ge\zeta\eta{\rbrace}\mathit{d_{uv}}},
\end{equation}
\begin{equation}
\mathcal{L}_{expan}={\sum_{j=0}^{N+M-1}}\mathcal{L}_{tree}(d_j,\varphi),
\end{equation}
where $\mathcal{T}(\cdot)$ is the minimal spanning tree based on $\mathcal{P}_c$, and $\mathit{d}_{uv}$ is the Euclidean distance between vertex $u$ and $u$ in the minimum tree. 
$\mathbb{I}$ is the indicator function, whereas $\eta$ is the average edge length of the minimal tree and $\mathbb{I}$. 
$\zeta$ is a scaling ratio used to modify the distance penalty. 
$d_j$ represents the expected compact points surrounding the $j$-th new hyperpoint.

Finally, we incorporate the trade-off indicator $\tau$ into the tightly constrained loss function, resulting in the formulation of the final loss function as follows:
\begin{equation}
\mathcal{L}=\mathcal{L}_{rec}+\tau\mathcal{L}_{expan},
\end{equation}

\section{Experiments}
In this section, we comprehensively evaluate our 3DMamabaComplete by conducting meticulous comparisons with various state-of-the-art techniques, including TopNet \cite{tchapmi2019topnet}, PCN \cite{yuan2018pcn}, FoldingNet \cite{yang2018foldingnet}, GRNet \cite{xie2020grnet}, ECG \cite{pan2020ecg}, CRN \cite{wang2020cascaded}, ASFM \cite{xia2021asfm}, PoinTr \cite{yu2021pointr}, SnowflakeNet \cite{xiang2021snowflakenet}, ProxyFormer \cite{li2023proxyformer}, and AnchorFormer \cite{chen2023anchorformer}. Our aim is to provide a rigorous and logically grounded analysis of their performance and capabilities.

\begin{table}[t] 
    \centering
    \caption{Quantitative comparisons for point cloud completion via CD-$\ell_1~\times\!10^3$ metric on PCN dataset \cite{yuan2018pcn}. (Lower is better, best scores are in \textbf{bold}.)}
    \resizebox{0.9\linewidth}{!}{%
    \begin{tabular}{c|c|cccccccc}
    \toprule
    Methods  & Avg.   & Airplane & Cabinet & Car    & Chair  & Lamp   & Sofa   & Table  & Watercraft \\   \midrule
    TopNet~\cite{tchapmi2019topnet}       & 15.499 & 9.314    & 17.923  & 13.736 & 17.952 & 17.324 & 18.667 & 15.221 & 13.854   \\
    PCN~\cite{yuan2018pcn}          & 11.697 & 6.834    & 12.719  & 9.923  & 13.588 & 13.954 & 14.608 & 11.402 & 10.545     \\
    FoldingNet~\cite{yang2018foldingnet}   & 12.231 & 8.542    & 13.252  & 11.202 & 13.901 & 13.618 & 14.218 & 11.935 & 11.183     \\
    GRNet~\cite{xie2020grnet}        & 12.809 & 8.647    & 14.904  & 11.602 & 14.469 & 13.379 & 15.374 & 13.125 & 10.971     \\
    ECG~\cite{pan2020ecg}          & 10.481 & 6.139    & 13.180   & 9.404  & 10.831 & 10.265 & 13.146 & 12.560  & 8.324      \\
    CRN~\cite{wang2020cascaded}           & 12.891 & 7.878    & 15.320   & 12.598 & 13.927 & 13.303 & 16.144 & 12.771 & 11.190      \\
    ASFM~\cite{xia2021asfm}         & 11.723 & 7.067    & 14.439  & 10.994 & 13.280  & 12.330  & 14.583 & 10.832 & 10.257     \\
    PoinTr~\cite{yu2021pointr}       & 9.006  & 5.343    & 11.033  & 9.118  & 10.137 & 8.388  & 11.762 & 8.534  & 7.732      \\ 
    SnowflakeNet~\cite{xiang2021snowflakenet}          & 8.362 & 5.262    & 10.372   & 8.847 & 9.103 & 7.717 & 10.714 & 7.663 & 7.221  \\ 
    ProxyFormer~\cite{li2023proxyformer}     & 11.342 &  6.346  & 12.373 & 9.564 & 13.159 & 13.395 & 14.340 & 11.346 &   10.212   \\ 
    AnchorFormer~\cite{chen2023anchorformer}    &  7.371 & 3.972     &  9.596   &  8.528 & 8.458 & 6.387 & 9.130 & 6.735  &   6.163    \\ 
    \textcolor{black}{CRA-PCN~\cite{rong2024cra}}  & 7.432 	& 3.978  & 9.692  & 8.523  & 8.525 	& 6.356  & 9.330  &6.820  & 6.231 \\
    \textcolor{black}{MMDR~\cite{fei2025multi}}   & 8.081 & 5.011 & 10.404 & 8.883 & 8.701 & 6.682 & 10.532 & 7.762 & 6.761\\
    \hline
    \textbf{3DMambaComplete}    & \textbf{6.907} & \textbf{3.861}  &  \textbf{9.115}   &\textbf{7.721}  & \textbf{7.412}  & \textbf{5.733}  & \textbf{9.042} & \textbf{6.288}  & \textbf{6.085}      \\ 
    \bottomrule
    \end{tabular} %
    }
    \label{tab:pcn_cd_l1_tomm}
\end{table}

\subsection{Dataset}
To validate the effectiveness of our proposed 3DMambaComplete approach, we conduct evaluations on three benchmark datasets.

\textbf{PCN dataset \cite{yuan2018pcn}}: 
The PCN dataset is a commonly used benchmark dataset in academia for point cloud completion tasks.
It is made up of eight categories with a total of 30,974 forms taken from the ShapeNet dataset. 
Each whole point cloud in this dataset has exactly 16,384 points since it was created using the backward projection technique from eight different perspectives.

\textbf{KITTI dataset \cite{geiger2012we}}: 
The KITTI dataset captures real-world traffic scenes using a combination of a 3D laser scanner, a high-resolution RGB camera, and a grayscale stereo camera. 
However, the original KITTI dataset lacks annotations and exhibits significant variations in the number of points in incomplete car shapes. 
Moreover, it lacks essential ground truth labels necessary for quantitative evaluation. 
In this study, we follow the standard settings proposed in \cite{yu2021pointr,zhou2022seedformer} and utilize the entire set of shapes in the KITTI dataset for testing purposes. 
Additionally, we fine-tune the trained model on a car-specific subset of PCN dataset that encompasses all car shapes.

\textbf{ShapeNet34/55datasets \cite{yu2021pointr}}: 
Existing datasets for point cloud completion often have limitations in terms of the variety of object shape classes and quantities, which may not adequately represent real-world scenes. In order to address this issue, the ShapeNet-55/34 dataset has been developed to provide a more comprehensive evaluation framework. This dataset comprises 55 clean point cloud shapes that represent common objects, enabling diverse performance assessment. The training set consists of 41,952 shapes, while the test set contains 10,518 shapes, making it a valuable resource for training and validating models.

Additionally, the ShapeNet34 benchmark enhances the categorization of the original ShapeNet dataset by dividing it into 34 visible and 21 invisible categories. 
The training set comprises 46,765 shapes from the visible category, whereas the test set consists of 3,400 shapes from the visible category and an additional 2,305 shapes from the invisible category.
This division increases the difficulty of the testing process and better reflects the complexity of real-world scenarios. 
It enables a more rigorous evaluation of the generalization ability of point cloud completion models.

\begin{table}[t]
\centering
\caption{Quantitative comparisons for point cloud completion via CD-$\ell_2~\times\!10^3$ metric on PCN dataset \cite{yuan2018pcn}. (Lower is better, best scores are in \textbf{bold}.)}
\resizebox{0.9\linewidth}{!}{%
\begin{tabular}{c|c|cccccccc}
\toprule
Methods & Avg.  & Airplane & Cabinet & Car & Chair & Lamp & Sofa & Table & Watercraft  \\ \midrule
ASFM~\cite{xia2021asfm} & 0.918 & 0.334 & 1.158 & 0.597 & 1.030 & 1.187 & 1.180 & 1.212 & 0.647 \\
CRN~\cite{wang2020cascaded} & 0.628 & 0.351 & 0.704 & 0.526 & 0.653 & 0.842 & 0.896 & 0.570 & 0.480  \\
ECG~\cite{pan2020ecg} & 0.408 & 0.169 & 0.502 & 0.253 & 0.459 & 0.570 & 0.632 & 0.380 & 0.299  \\
FoldingNet~\cite{yang2018foldingnet} & 0.570 & 0.264 & 0.665 & 0.355 & 0.719 & 0.733 & 0.687 & 0.624 & 0.514  \\
GRNet~\cite{xie2020grnet} & 0.662 & 0.405 & 0.786 & 0.462 & 0.808 & 0.689 & 0.950 & 0.672 & 0.525  \\
PCN~\cite{yuan2018pcn} & 0.542 & 0.213 & 0.633 & 0.313 & 0.638 & 0.750 & 0.784 & 0.573 & 0.433  \\
TopNet~\cite{tchapmi2019topnet} & 0.599  & 0.240 & 0.712 & 0.404 & 0.856 & 0.710 & 0.771 & 0.595 & 0.504 \\ 
PoinTr~\cite{yu2021pointr} & 0.564  & 0.342 & 0.682 & 0.415 & 0.704 & 0.616 & 0.815 & 0.526 & 0.415  \\
Snowflake~\cite{xiang2021snowflakenet} & 0.311 & 0.131 & 0.439 & 0.246 & 0.343 & 0.343 & 0.493 & 0.294 & 0.196  \\ 
ProxyFormer~\cite{li2023proxyformer} & 0.565 & 0.190 & 0.562 & 0.307 & 0.661 & 0.857 & 0.864 & 0.651 & 0.427 \\ 
AnchorFormer~\cite{chen2023anchorformer} & 0.224  & 0.064  & 0.348 & 0.223 & 0.286 & 0.213 & 0.315 & 0.211 & \textbf{0.135}  \\
\textcolor{black}{CRA-PCN~\cite{rong2024cra}}  & 0.223 & \textbf{0.063} & 0.328 & 0.204  & 0.259  & 0.210 & \textbf{0.310} & 0.208 & 0.199   \\
\textcolor{black}{MMDR~\cite{fei2025multi}}  & 0.242 & 0.076 & 0.367 & 0.221 & 0.255 	& 0.193 & 0.358 & 0.256 & 0.207  	   \\
\midrule
\textbf{3DMambaComplete} &\textbf{0.216}  &0.067  &\textbf{0.325}  &\textbf{0.203}  &\textbf{0.254}  &\textbf{0.182}  &0.357  &\textbf{0.188}  &0.153    \\ \bottomrule
\end{tabular}%
}
\label{tab:pcn_cd_l2_tomm}
\end{table}

\begin{figure*}[htbp]
    \centering
    \includegraphics[width=\linewidth]{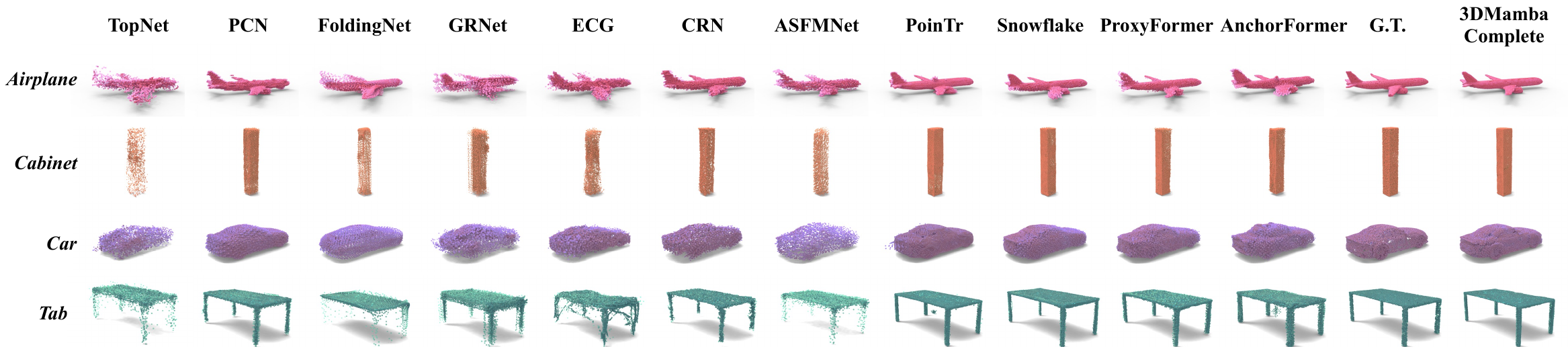}
    \caption{Comparison of different point cloud completion visualization on the PCN dataset \cite{yuan2018pcn}.}
    \label{Figure:PCN_comparison}
\end{figure*}

\subsection{Implementation Details}
More information on the 3DMambaComplete implementation is given in this section.
Firstly, we employ FPS \cite{qi2017pointnet} and EdgeConv for sampling the input point cloud, resulting in the acquisition of $\mathcal{P}_0$ comprising 128 points and their corresponding features $\mathcal{F}_0$.
Subsequently, we input them into the Mamba Encoder, utilizing pooling operations to extract hyperpoint features and employing multi-head cross-attention to determine their respective coordinate values. Each Cross-Attention Block predicts a set of hyperpoints, where M represents the number of hyperpoints (set to 16), and L represents the total number of hyperpoints (set to 128).
Afterwards, we decode the sampled and hyper points. The decoder in the point deformation stage consists of six layers of Mamba Blocks and MLPs. The grid point number K is set to 64. The parameters $\alpha$ and $\lambda$ are determined through cross-validation, with empirical values of 1.2 and 0.05, respectively. The network is trained using the AdamW optimizer with a base learning rate of 0.0002.
To enhance data diversity in both the training and testing phases, we randomly select the value of `n' from 2,048, 4,096, or 6,144, representing 25$\%$, 50$\%$, or 75$\%$ of the complete point cloud, respectively. These values correspond to the \emph{Simple}, \emph{Medium}, and \emph{Hard} difficulty levels.

\subsection{Experimental Metrics}

To evaluate the performance of the synthetic datasets ShapeNet55/34 and PCN, we utilize the $\mathit{L}_1$ and $\mathit{L}_2$ Chamfer Distance as well as the F-score as evaluation metrics.
The Chamfer Distance is a metric that calculates the average distance between the nearest points in two point clouds, namely the forecast point cloud $P$  and the ground truth $G$ \cite{nguyen2021point}. 
Mathematically, the Chamfer Distance can be defined as follows:
\begin{equation}
CD(\boldsymbol{P},\boldsymbol{G})={\dfrac{1}{|\boldsymbol{P}|}}{\sum_{p{\in}\boldsymbol{P}}}{\underset{g{\in}\boldsymbol{G}}{min}}{\lVert}p-g{\rVert}+{\dfrac{1}{|\boldsymbol{G}|}}{\sum_{g{\in}\boldsymbol{G}}}{\underset{p{\in}\boldsymbol{P}}{min}}{\lVert}g-p{\rVert},
\end{equation}

To assess the accuracy of point cloud completion, the F-score metric combines the reconciled values of precision and recall \cite{sokolova2006beyond}.
Due to the absence of ground truth values in the KITTI dataset, the evaluation of performance necessitates the utilization of the MMD and FD metrics. 
MMD (Minimum Matching Distance) \cite{tesema2023point} operates as a local measure, evaluating the precision of correspondences between two point clouds through the computation of distances among corresponding points. 
On the other hand, FD (Fidelity Error) \cite{fei2022comprehensive} serves as a metric to assess dissimilarities between reconstructed and original point clouds by averaging the distances between the closest points in both clouds. The calculation method is as follows:
\begin{equation}
FD\left(S_{1}, S_{2}\right)=\frac{1}{\left|S_{1}\right|} \sum_{x \in S_{1}} \min _{y \in S_{2}}\|x-y\|_{2},
\end{equation}

\subsection{Experiments Performances}

\begin{table}[t]
\centering
\caption{Quantitative comparisons for point cloud completion via F-score$@1\%$ metric on PCN dataset \cite{yuan2018pcn}. (Higher is better, best scores are in \textbf{bold}.)}
\resizebox{0.9\linewidth}{!}{%
    \begin{tabular}{c|c|cccccccc}
    \toprule
    Methods  & Avg. & Airplane & Cabinet & Car & Chair & Lamp & Sofa & Table & Watercraft \\ \midrule
    ASFM~\cite{xia2021asfm} & 0.459 & 0.738 & 0.330 & 0.409 & 0.399 & 0.496 & 0.331 & 0.411 & 0.556  \\
    CRN~\cite{wang2020cascaded} & 0.549 & 0.804 & 0.439 & 0.486 & 0.505 & 0.533 & 0.409 & 0.606 & 0.607  \\
    ECG~\cite{pan2020ecg} & 0.684 & 0.870 & 0.605 & 0.675 & 0.635 & 0.698 & 0.502 & 0.730 & 0.759  \\
    FoldingNet~\cite{yang2018foldingnet} & 0.418 & 0.723 & 0.310 & 0.488 & 0.307 & 0.332 & 0.280 & 0.457 & 0.449  \\
    GRNet~\cite{xie2020grnet} & 0.541  & 0.711 & 0.447 & 0.507 & 0.483 & 0.609 & 0.415 & 0.556 & 0.603 \\
    PCN~\cite{yuan2018pcn} & 0.589 & 0.831 & 0.508 & 0.649 & 0.506 & 0.522 & 0.433 & 0.621 & 0.644  \\
    TopNet~\cite{tchapmi2019topnet} & 0.443 & 0.760 & 0.321 & 0.496 & 0.342 & 0.345 & 0.313 & 0.504 & 0.461 \\ 
    PoinTr~\cite{yu2021pointr} & 0.527 & 0.704 & 0.435 & 0.549 & 0.447 & 0.528 & 0.384 & 0.602 & 0.570  \\
    Snowflake~\cite{xiang2021snowflakenet}& 0.743 & 0.897 & 0.648 & 0.697 & 0.705 & 0.790 & 0.604 & 0.820 & 0.787  \\ 
    ProxyFormer~\cite{li2023proxyformer} & 0.615 & 0.847 & 0.566 & 0.650 & 0.522 & 0.544 & 0.467 & 0.669 & 0.655  \\ 
    AnchorFormer~\cite{chen2023anchorformer} & 0.790 & 0.952 & 0.676 & 0.717 & 0.743 & 0.849 & 0.678 & 0.854 & 0.849  \\ 
    \textcolor{black}{CRA-PCN~\cite{rong2024cra}}  & 0.795 & \textbf{0.961} & 0.682 & 0.721 & 0.745 & 0.856 & 0.683 & 0.861 & \textbf{0.850}   \\
    \textcolor{black}{MMDR~\cite{fei2025multi}}  & 0.784 & 0.949 & 0.661 & 0.697 & 0.740 & 0.851 & 0.677 & 0.854 & 0.841   \\
    \midrule
    \textbf{3DMambaComplete} &\textbf{0.805} & 0.951  &\textbf{0.694}  &\textbf{0.759}  &\textbf{0.778}  &\textbf{0.867}  &\textbf{0.689}    &\textbf{0.859}  & 0.841    \\ \bottomrule
    \end{tabular}%
    }
\label{tab:pcn_f1_score_tomm}
\end{table}

\subsubsection{Results on PCN Dataset}

The comparative results for the eight categories in the PCN dataset are provided in Tables \ref{tab:pcn_cd_l1_tomm}, \ref{tab:pcn_cd_l2_tomm}, and \ref{tab:pcn_f1_score_tomm}. The experimental data consistently demonstrate the outstanding performance of the proposed 3DMambaComplete model across all categories, significantly outperforming other baseline methods.
Specifically, in terms of the CD-$\ell_1$ metric, 3DMambaComplete achieves the best results in all eight categories as well as the mean value. For the CD-$\ell_2$ metric, it obtains the highest average score and ranks first in five categories. Regarding the F-score$@1\%$, the model also leads in the average value and achieves top performance in six categories.
Notably, 3DMambaComplete comprehensively surpasses its main competitor, Anchorformer, across all categories, with remarkable average scores of 6.907, 0.216, and 0.805 for CD-${\ell}_1$, CD-${\ell}_2$, and F-score$@1\%$, respectively, demonstrating consistent and superior performance.
Furthermore, the visual comparison of point cloud completions by 3DMambaComplete in Figure~\ref{Figure:PCN_comparison} intuitively validates the effectiveness of the proposed method in processing incomplete point clouds. By incorporating Mamba blocks, the model is capable of effectively handling long sequences, comprehensively capturing information from the input point cloud, and generating hyperpoints with rich features, ultimately producing complete point clouds with detailed structural information.

\subsubsection{Results on KITTI Dataset}

\begin{figure}[htbp]
    \centering
    \includegraphics[width=0.9\linewidth]{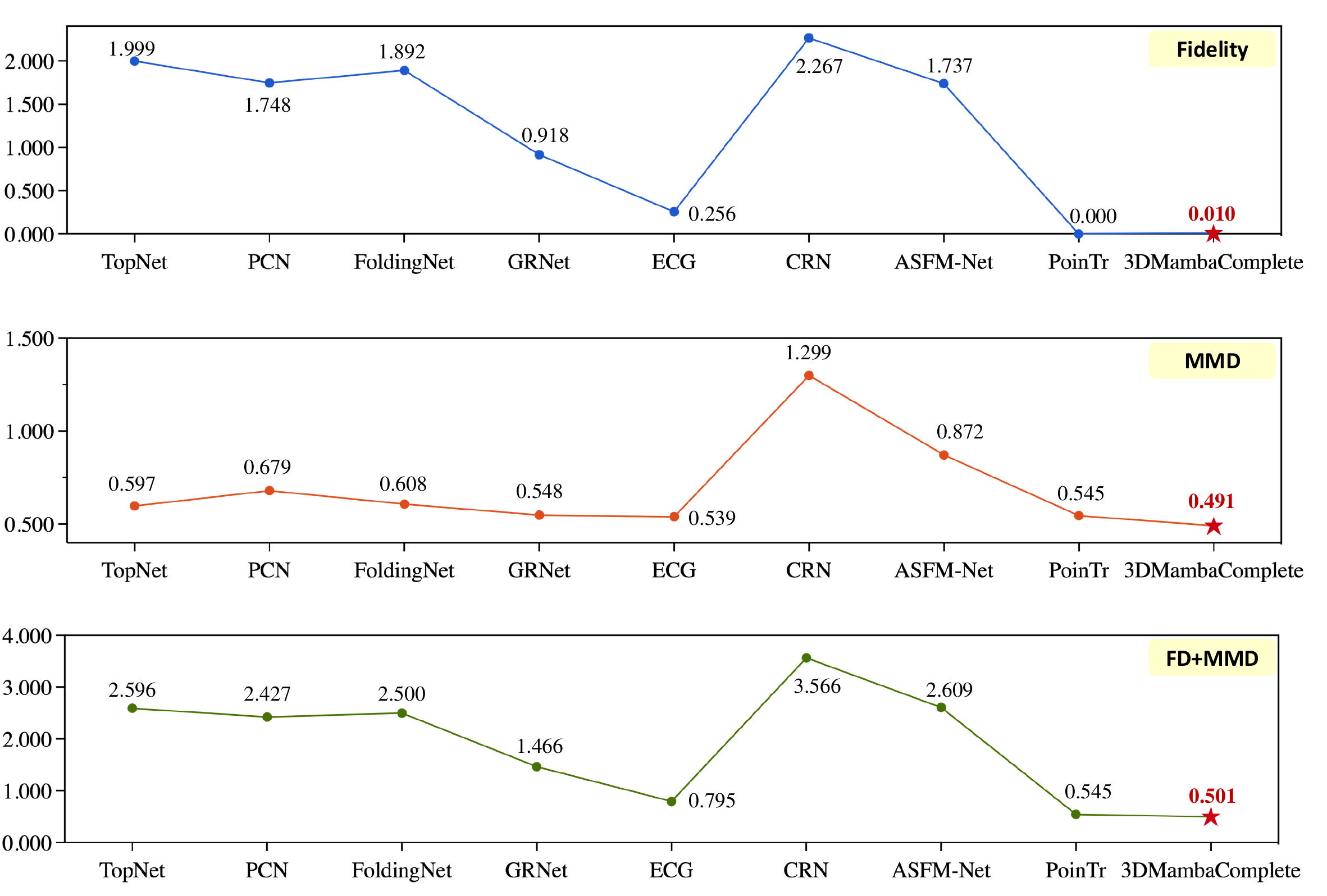}
    \caption{Quantitative comparison on KITTI \cite{geiger2013vision} dataset with metrics of MMD and FD.}
    \label{fig:kitti_results}
\end{figure}

\begin{figure}[htbp]
    \centering
    \includegraphics[width=\linewidth]{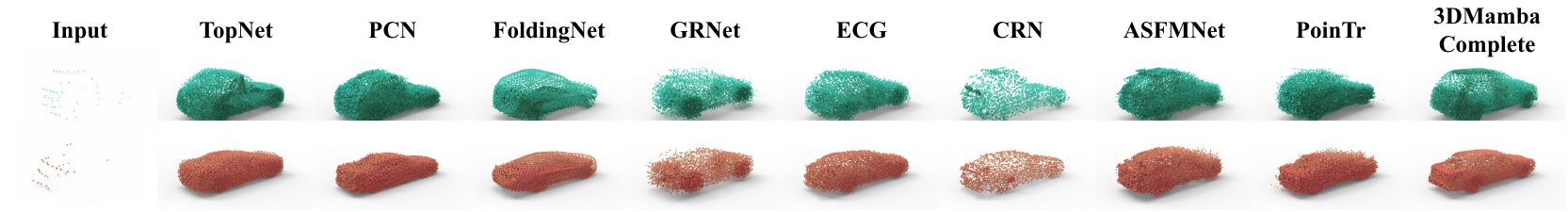}
    \caption{Comparison of different point cloud completion visualization on the KITTI \cite{geiger2013vision} dataset.}
    \label{Figure:KITTI_comparison}
\end{figure}

We fine-tuned the model on the PCN dataset, followed by pre-training on the KITTI dataset. Figure \ref{fig:kitti_results} presents a comprehensive comparison of 3DMambaComplete with other state-of-the-art methods on the KITTI dataset. The results indicate that 3DMambaComplete achieves the best performance on both MMD and FD metrics, while obtaining sub-optimal results on the Fidelity metric. Additionally, Figure \ref{Figure:KITTI_comparison} illustrates the visual comparison on the KITTI dataset.
As demonstrated in this figure, 3DMambaComplete exhibits superior performance when addressing real scanned data characterized by significant missing input information. The reconstructed point cloud structures maintain a high level of fine detail, even when the input point cloud data is sparse. This observation underscores the capability of 3DMambaComplete to accurately capture the features of input point clouds through the generation of hyperpoints and the optimization of point sampling.

\begin{table}[htbp]
  \centering
  \caption{Quantitative Comparison of CD-$\ell_1$/CD-$\ell_2$ $\times 10^{-3}$ and average F-Score@1\% metrics for different point cloud completion methods on ShapeNet55 \cite{yu2021pointr}. (Best scores are in \textbf{bold}.)}
  \resizebox{0.9\linewidth}{!}{%
    \begin{tabular}{c|ccccc}
    \toprule
    \cellcolor{white} Methods &
      \begin{tabular}[c]{@{}c@{}}CD-S\\ (CD-$\ell_1$/CD-$\ell_2$)\end{tabular} &
      \begin{tabular}[c]{@{}c@{}}CD-M\\ (CD-$\ell_1$/CD-$\ell_2$)\end{tabular} &
      \begin{tabular}[c]{@{}c@{}}CD-H\\ (CD-$\ell_1$/CD-$\ell_2$)\end{tabular} &
      \begin{tabular}[c]{@{}c@{}}CD-Avg\\ (CD-$\ell_1$/CD-$\ell_2$)\end{tabular} &
      \begin{tabular}[c]{@{}c@{}}F-score@1\%\\ Avg\end{tabular} \\ \midrule
    TopNet \cite{tchapmi2019topnet}           & 27.233 / 2.483          & 28.749 / 2.848 & 33.986 / 4.642 & 29.989 / 3.324 & 0.110          \\
    PCN \cite{yuan2018pcn}                    & 22.990 / 1.811          & 23.976 / 2.062 & 27.360 / 2.937 & 24.775 / 2.270 & 0.167          \\
    FoldingNet \cite{yang2018foldingnet}      & 25.203 / 2.095          & 26.596 / 2.410 & 30.424 / 3.330 & 27.407 / 2.612 & 0.091          \\
    GRNet \cite{xie2020grnet}                 & 19.157 / 1.137          & 20.647 / 1.489 & 24.037 / 2.394 & 21.280 / 1.673 & 0.239          \\
    ECG \cite{pan2020ecg}                     & 16.709 / 1.167          & 18.726 / 1.545 & 23.478 / 2.555 & 19.638 / 1.756 & 0.321          \\
    CRN \cite{wang2020cascaded}               & 21.207 / 1.502          & 22.364 / 1.801 & 25.849 / 2.726 & 23.140 / 2.010 & 0.205          \\
    ASFM-Net \cite{xia2021asfm}               & 19.136 / 1.307          & 20.170 / 1.517 & 23.512 / 2.282 & 20.940 / 1.702 & 0.247          \\
    PoinTr \cite{yu2021pointr}                & 12.491 / 0.698 & 14.181 / 1.049 & 18.812 / 2.022 & 15.161 / 1.256 & \textbf{0.446} \\
    SnowflakeNet \cite{xiang2021snowflakenet} & 14.596 / 0.817          & 16.644 / 1.175 & 21.545 / 2.207 & 17.595 / 1.406 & 0.343          \\
    ProxyFormer \cite{li2023proxyformer}      & 21.463 / 1.637          & 22.226 / 1.827 & 25.114 / 2.591 & 22.934 / 2.018 & 0.198          \\
    AnchorFormer \cite{chen2023anchorformer}  & 14.662 / 0.853          & 15.504 / 1.087 & 18.096 / 1.770 & 16.087 / 1.237 & 0.328          \\ 
    \textcolor{black}{CRA-PCN~\cite{rong2024cra}}   & 11.987 / 0.638     & 13.454 / 0.961 & 16.982 / 1.799   & 14.141 / 1.133  & 0.399 \\
    \textcolor{black}{MMDR~\cite{fei2025multi}}  & \textbf{11.534} / 0.643          & \textbf{13.281} / 0.978   & 17.245 / 1.919 & 14.020 / 1.180 & 0.355   \\
    \midrule
    \textbf{3DMambaComplete} &
      12.698 / \textbf{0.617} &
      13.431 / \textbf{0.775} &
      \textbf{15.384} / \textbf{1.194} &
      \textbf{13.837} / \textbf{0.862} &
      0.341 \\ 
      \bottomrule
    \end{tabular}
    }
    \label{tab:shapnet55_tomm}%
\end{table}
\begin{table}[htbp]
  \centering
  \caption{Quantitative comparisons for point cloud completion via CD-$\ell_1~\times\!10^3$ metric on ShapeNet55 dataset \cite{yu2021pointr}. (Lower is better, best scores are in \textbf{bold}.)}
  \resizebox{0.95\linewidth}{!}{%
    \begin{tabular}{c|cccccccccc}
    \toprule
    Methods & Table & Chair & Airplane & Car   & Sofa  & Bridhouse & Bag   & Remote & Keyboard & Rocket \\
    \midrule
    TopNet \cite{tchapmi2019topnet} & 25.106  & 29.754  & 20.034  & 27.212  & 28.739  & 38.762  & 30.489  & 25.026  & 18.306  & 19.874  \\
    PCN \cite{yuan2018pcn} & 21.694  & 24.558  & 16.428  & 23.398  & 24.364  & 34.575  & 26.025  & 18.418  & 16.272  & 18.590  \\
    FoldingNet \cite{yang2018foldingnet} & 23.298  & 27.193  & 19.749  & 25.383  & 26.753  & 36.924  & 27.559  & 20.136  & 18.123  & 19.423  \\
    GRNet \cite{xie2020grnet} & 19.397  & 21.213  & 14.830  & 21.931  & 22.026  & 27.352  & 22.028  & 17.277  & 15.678  & 13.596  \\
    ECG \cite{pan2020ecg} & 17.912  & 18.233  & 12.347  & 18.161  & 19.813  & 25.926  & 20.013  & 15.865  & 12.887  & 11.742  \\
    CRN \cite{wang2020cascaded} & 20.313  & 22.931  & 15.231  & 23.804  & 23.367  & 32.237  & 24.139  & 18.163  & 15.767  & 14.522  \\
    ASFM-Net \cite{xia2021asfm} & 17.788  & 21.216  & 14.242  & 21.076  & 20.875  & 27.936  & 21.092  & 16.140  & 14.256  & 9.894  \\
    PoinTr \cite{yu2021pointr} & 13.041  & 14.838  & 10.137  & 15.747  & 14.790  & 20.670  & 15.295  & 13.722  & 10.159  & 9.842  \\
    Snowflake \cite{xiang2021snowflakenet} & 15.194  & 16.918  & 12.394  & 18.571  & 16.902  & 22.415  & 17.401  & 13.618  & 11.602  & 11.684  \\
    ProxyFormer \cite{li2023proxyformer} & 21.321  & 23.549  & 16.181  & 24.002  & 22.993  & 33.334  & 24.283  & 18.064  & 16.167  & 16.341  \\
    AnchorFormer \cite{chen2023anchorformer} & 14.040  & 15.768  & 10.969  & 18.145  & 15.971  & 22.742  & 15.599  & 12.167  & 11.109  & 10.543  \\
    \textcolor{black}{CRA-PCN~\cite{rong2024cra}}   & 13.516 & 14.791 & 10.101 & 17.231 	& 14.902 	& 19.976 & 14.828 & 11.758 	& 10.406 & 9.890 \\
    \textcolor{black}{MMDR~\cite{fei2025multi}}  & \textbf{12.449} & \textbf{13.482} & \textbf{9.529} & \textbf{15.712} & \textbf{13.653} & \textbf{18.178} & \textbf{13.531} & \textbf{10.577} & \textbf{9.570} & \textbf{8.699} \\
    \midrule
    \textbf{3DMambaComplete} & 13.489 & 14.781 & 10.033 & 17.150 & 15.347 & 20.798 & 14.815 & 11.723 & 10.612 & 9.627 \\
    \bottomrule
    \end{tabular}%
    }
  \label{tab:shapenet55_appendix_cd_l1}%
\end{table}%

\begin{table}[htbp]
  \centering
  \caption{Quantitative comparisons for point cloud completion via CD-$\ell_2~\times\!10^3$ metric on ShapeNet55 dataset \cite{yu2021pointr}. (Lower is better, best scores are in \textbf{bold}.)}
  \resizebox{0.95\linewidth}{!}{%
    \begin{tabular}{c|cccccccccc}
    \toprule
    Methods & Table & Chair & Airplane & Car   & Sofa  & Bridhouse & Bag   & Remote & Keyboard & Rocket \\
    \midrule
    TopNet \cite{tchapmi2019topnet} & 2.438 & 2.950  & 1.397 & 2.089 & 2.449 & 4.899 & 2.998 & 1.982 & 1.091 & 1.498 \\
    PCN \cite{yuan2018pcn} & 1.812 & 2.084 & 0.985 & 1.531 & 1.765 & 4.022 & 2.293 & 1.042 & 0.924 & 1.318 \\
    FoldingNet \cite{yang2018foldingnet} & 1.953 & 2.350  & 1.331 & 1.777 & 2.076 & 4.291 & 2.439 & 1.209 & 1.054 & 1.260 \\
    GRNet \cite{xie2020grnet} & 1.355 & 1.575 & 0.830  & 1.351 & 1.435 & 7.735 & 1.691 & 0.953 & 0.783 & 0.855 \\
    ECG \cite{pan2020ecg} & 1.543 & 1.409 & 0.657 & 1.033 & 1.421 & 2.830  & 1.677 & 0.961 & 0.688 & 0.701 \\
    CRN \cite{wang2020cascaded} & 1.570  & 1.859 & 0.842 & 1.606 & 1.663 & 3.599 & 2.040  & 1.041 & 0.815 & 0.930 \\
    ASFM-Net \cite{xia2021asfm} & 1.241 & 1.595 & 0.748 & 1.293 & 1.371 & 2.769 & 1.618 & 0.880  & 0.677 & 0.867 \\
    PoinTr \cite{yu2021pointr} & 0.979 & 1.149 & 0.547 & \textbf{0.974} & 0.944 & 2.131 & 1.179 & 0.992 & 0.452 & 0.627 \\
    Snowflake \cite{xiang2021snowflakenet} & 1.053 & 1.267 & 0.677 & 1.181 & 1.053 & 2.162 & 1.325 & 0.714 & 0.494 & 0.722 \\
    ProxyFormer \cite{li2023proxyformer} & 1.934 & 2.003 & 0.978 & 1.630  & 1.682 & 3.837 & 2.121 & 1.037 & 0.906 & 1.137 \\
    AnchorFormer \cite{chen2023anchorformer} & 0.989 & 1.164 & 0.548 & 1.158 & 1.021 & 2.523 & 1.118 & 0.591 & 0.592 & 0.642 \\
    \textcolor{black}{CRA-PCN~\cite{rong2024cra}}   & 0.942 & 1.201 & 0.536 & 1.142 & 0.919 	& 2.131 & 1.123 & 0.575 & 0.488 & 0.589 \\
    \textcolor{black}{MMDR~\cite{fei2025multi}}   & 0.941 & 1.041 & 0.529 & 1.050 & 0.899 & \textbf{1.821} & 1.015 & 0.558 & 0.476 & 0.558 \\
    \midrule
    \textbf{3DMambaComplete} & \textbf{0.854} & \textbf{0.972} & \textbf{0.468} & 1.029 & \textbf{0.894} & 1.934 & \textbf{0.946} & \textbf{0.552} & \textbf{0.416} & \textbf{0.551} \\
    \bottomrule
    \end{tabular}%
    }
  \label{tab:shapenet55_appendix_cd_l2}%
\end{table}%

\begin{table}[htbp]
  \centering
  \caption{Quantitative comparisons for point cloud completion via F-score$@1\%$ metric on ShapeNet55 dataset \cite{yu2021pointr}. (Higher is better, best scores are in \textbf{bold}.)}
  \resizebox{0.95\linewidth}{!}{%
    \begin{tabular}{c|cccccccccc}
    \toprule
    Methods & Table & Chair & Airplane & Car   & Sofa  & Bridhouse & Bag   & Remote & Keyboard & Rocket \\
    \midrule
    TopNet \cite{tchapmi2019topnet} & 0.147  & 0.088  & 0.203  & 0.077  & 0.077  & 0.046  & 0.084  & 0.121  & 0.218  & 0.278  \\
    PCN \cite{yuan2018pcn} & 0.181  & 0.143  & 0.340  & 0.103  & 0.101  & 0.062  & 0.127  & 0.228  & 0.272  & 0.343  \\
    FoldingNet \cite{yang2018foldingnet} & 0.163  & 0.066  & 0.172  & 0.063  & 0.061  & 0.027  & 0.066  & 0.139  & 0.201  & 0.209  \\
    GRNet \cite{xie2020grnet} & 0.231  & 0.220  & 0.412  & 0.150  & 0.167  & 0.135  & 0.210  & 0.306  & 0.293  & 0.536  \\
    ECG \cite{pan2020ecg} & 0.320  & 0.336  & 0.559  & 0.255  & 0.234  & 0.194  & 0.285  & 0.374  & 0.454  & 0.639  \\
    CRN \cite{wang2020cascaded} & 0.214  & 0.183  & 0.357  & 0.118  & 0.142  & 0.102  & 0.163  & 0.247  & 0.266  & 0.485  \\
    ASFM-Net \cite{xia2021asfm} & 0.294  & 0.209  & 0.426  & 0.164  & 0.177  & 0.122  & 0.219  & 0.332  & 0.358  & 0.522  \\
    PoinTr \cite{yu2021pointr} & \textbf{0.480}  & \textbf{0.438}  & \textbf{0.598}  & \textbf{0.368}  & \textbf{0.394}  & \textbf{0.348}  & \textbf{0.416}  & \textbf{0.467}  & \textbf{0.533}  & \textbf{0.693}  \\
    Snowflake \cite{xiang2021snowflakenet} & 0.373  & 0.345  & 0.497  & 0.222  & 0.295  & 0.233  & 0.336  & 0.426  & 0.456  & 0.611  \\
    ProxyFormer \cite{li2023proxyformer} & 0.195  & 0.156  & 0.346  & 0.096  & 0.127  & 0.070  & 0.137  & 0.230  & 0.255  & 0.433  \\
    AnchorFormer \cite{chen2023anchorformer} & 0.372  & 0.324  & 0.510  & 0.190  & 0.276  & 0.189  & 0.315  & 0.434  & 0.434  & 0.629  \\
    \textcolor{black}{CRA-PCN~\cite{rong2024cra}}  & 0.389 & 0.355 & 0.519 & 0.214 & 0.310 & 0.257 & 0.336 & 0.445 & 0.438 & 0.646  \\
    \textcolor{black}{MMDR~\cite{fei2025multi}}   & 0.369 & 0.351 & 0.513 & 0.228 & 0.306 & 0.254  & 0.335 & 0.423 & 0.426 & 0.648 \\
    \midrule
    \textbf{3DMambaComplete} & 0.326  & 0.299  & 0.513  & 0.166  & 0.240  & 0.177  & 0.280  & 0.389  & 0.375  & 0.649  \\
    \bottomrule
    \end{tabular}%
    }
  \label{tab:shapenet55_appendix_f1}%
\end{table}%

\subsubsection{Results on ShapeNet55 Dataset}

\begin{figure*}[ht]
    \centering
    \includegraphics[width=\linewidth]{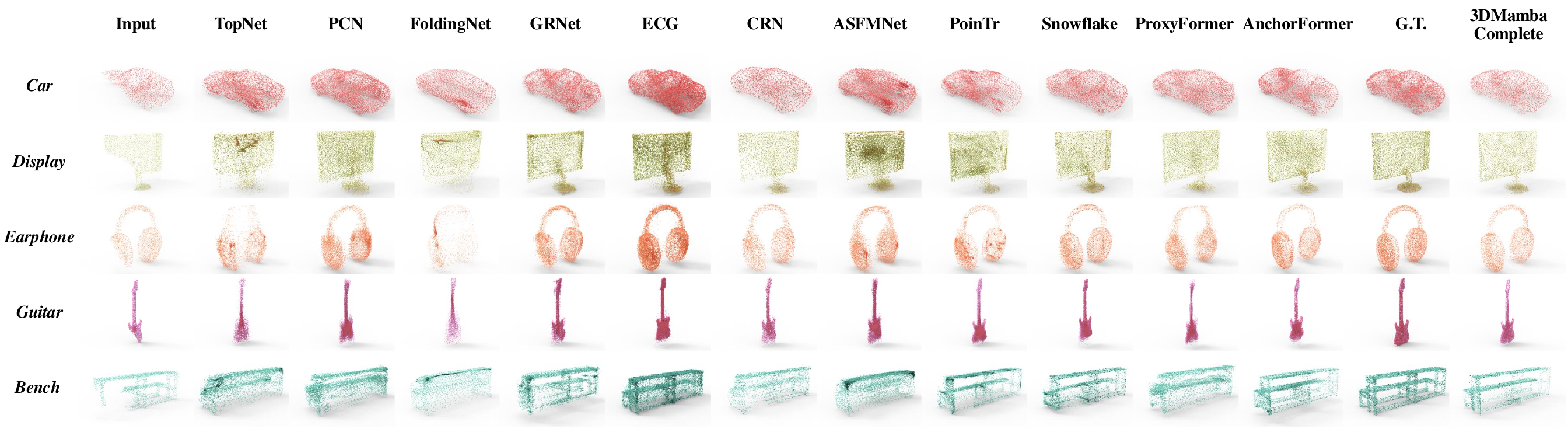}
    \caption{Comparison of different point cloud completion visualization on the ShapeNet55 Dataset \cite{yu2021pointr}.}
    \label{Figure:SN55_comparison}
\end{figure*}

Given the exceptional performance of 3DMambaComplete on the PCN and KITTI datasets, we assess its scalability to the ShapeNet55 dataset, which encompasses a wider range of object categories.
The results presented in Table \ref{tab:shapnet55_tomm} highlight the impressive capabilities of 3DMambaComplete across 55 distinct object categories, with average CD-${\ell}_1$, CD-${\ell}_2$, and F-score@1$\%$ values of 13.837, 0.862, and 0.341, respectively.
In particular, when the number of input point clouds `n' is equal to 2048, indicating a higher difficulty level denoted as ``\emph{H}'', 3DMambaComplete achieves state-of-the-art performance across all categories. It achieves an average CD-${\ell}_1$ of 15.384 and an average CD-${\ell}_2$ of 1.194.

Furthermore, Tables \ref{tab:shapenet55_appendix_cd_l1}, \ref{tab:shapenet55_appendix_cd_l2}, and \ref{tab:shapenet55_appendix_f1} provide a comprehensive comparison of metric results between five categories with abundant samples (the five left columns) and five categories with limited samples (the five right columns). Remarkably, 3DMambaComplete underperforms MMDR in the CD-${\ell}_1$  metric, it outperforms MMDR in the CD-${\ell}_2$  metric. This is because MMDR over-optimizes local details, which compromises the overall performance of object reconstruction. In contrast, 3DMambaComplete leverages the selective attention mechanism of Mamba to capture relevant information more effectively, resulting in higher-quality point cloud generation.
For F-score@1$\%$ metric, even though PointTr outperforms 3DMambaComplete, demonstrating its advantages in overall shape coverage and surface integrity, its CD-L1 and CD-L2 results are poor. This indicates that the method has shortcomings in precise point position reconstruction and fine-structure consistency.
The effectiveness of 3DMambaComplete is further corroborated through a visual comparison depicted in Figure \ref{Figure:SN55_comparison}. The results clearly illustrate that 3DMambaComplete surpasses other methods by generating highly accurate and detailed complete shapes across various environments.

\begin{table}[htbp]
  \centering
  \caption{Quantitative Comparison of CD-$\ell_1$/CD-$\ell_2$ $\times 10^3$ and average F-Score@$1\%$ metrics for different point cloud completion methods on the dataset ShapeNet34 \cite{yu2021pointr}.(Best scores are in \textbf{bold}.)}
    \resizebox{0.9\linewidth}{!}{%
    \begin{tabular}{c|ccccc}
    \toprule
      Methods & 
      \begin{tabular}[c]{@{}c@{}}CD-S\\      (CD-$\ell_1$/CD-$\ell_2$)\end{tabular} &
      \begin{tabular}[c]{@{}c@{}}CD-M\\      (CD-$\ell_1$/CD-$\ell_2$)\end{tabular} &
      \begin{tabular}[c]{@{}c@{}}CD-H\\      (CD-$\ell_1$/CD-$\ell_2$)\end{tabular} &
      \begin{tabular}[c]{@{}c@{}}CD-Avg\\      (CD-$\ell_1$/CD-$\ell_2$)\end{tabular} &
      \begin{tabular}[c]{@{}c@{}}F-score@$1\%$\\ Avg\end{tabular} \\
    \midrule
    TopNet \cite{tchapmi2019topnet} & 22.382 / 1.606 & 23.271 / 1.793 & 26.020 / 2.432 & 23.891 / 1.944 & 0.154 \\
    PCN \cite{yuan2018pcn}  & 21.433 / 1.551 & 22.304 / 1.753 & 25.086 / 2.426 & 22.941 / 1.910 & 0.192 \\
    FoldingNet \cite{yang2018foldingnet} & 23.556 / 1.859 & 24.466 / 2.059 & 28.584 / 2.759 & 25.535 / 2.226 & 0.137 \\
    GRNet \cite{xie2020grnet} & 18.809 / 1.102 & 20.032 / 1.365 & 22.990 / 2.090 & 20.610 / 1.519 & 0.247 \\
    ECG \cite{pan2020ecg}  & 13.123 / 0.735 & 14.627 / 0.996 & 18.459 / 1.696 & 15.403 / 1.142 & \textbf{0.496} \\
    CRN \cite{wang2020cascaded} & 20.304 / 1.362 & 21.216 / 1.594 & 24.159 / 2.318 & 21.893 / 1.758 & 0.221 \\
    ASFM-Net \cite{xia2021asfm} & 18.351 / 1.189 & 19.123 / 1.342 & 21.914 / 1.909 & 19.796 / 1.480 & 0.268 \\
    PoinTr \cite{yu2021pointr} & \textbf{12.006} / \textbf{0.632} & \textbf{13.393} / 0.910 & 17.364 / 1.697 & \textbf{14.254} / 1.080 & 0.459 \\
    SnowflakeNet \cite{xiang2021snowflakenet} & 13.966 / 0.714 & 15.613 / 0.987 & 19.646 / 1.730 & 16.408 / 1.144 & 0.362 \\
    ProxyFormer \cite{li2023proxyformer} & 21.463 / 1.637 & 22.226 / 1.827 & 25.114 / 2.591 & 22.934 / 2.018 & 0.198 \\
    AnchorFormer \cite{chen2023anchorformer} & 14.662 / 0.853 & 15.504 / 1.087 & 18.096 / 1.770 & 16.087 / 1.237 & 0.328 \\
    \textcolor{black}{CRA-PCN~\cite{rong2024cra}}   & 12.443 / 0.667  & 13.772 / 0.859  & 16.971 / 1.449  & 14.395 / 0.992  & 0.352 \\
    \textcolor{black}{MMDR~\cite{fei2025multi}}  & 11.571 / 0.654 & 13.111 / 0.938 & 16.738 / 1.769 & 13.807 / 1.120 & 0.362     \\
    \midrule
    \textbf{3DMambaComplete} & 13.247 / 0.661 & 14.056 / \textbf{0.844} & \textbf{16.385} / \textbf{1.391} & 14.563 / \textbf{0.965} & 0.324 \\
    \bottomrule
    \end{tabular}%
    }
  \label{tab:shapnet34_tomm}%
\end{table}%

\begin{table}[htbp]
  \centering
  \caption{Quantitative Comparison of CD-$\ell_1$/CD-$\ell_2$ $\times 10^3$ and average F-Score@$1\%$ metrics for different point cloud completion methods on the dataset ShapeNetUnseen21 \cite{yu2021pointr}. (Best scores are in \textbf{bold}.)}
    \resizebox{0.9\linewidth}{!}{%
    \begin{tabular}{c|ccccc}
    \toprule
      Methods & 
      \begin{tabular}[c]{@{}c@{}}CD-S\\      (CD-$\ell_1$/CD-$\ell_2$)\end{tabular} &
      \begin{tabular}[c]{@{}c@{}}CD-M\\      (CD-$\ell_1$/CD-$\ell_2$)\end{tabular} &
      \begin{tabular}[c]{@{}c@{}}CD-H\\      (CD-$\ell_1$/CD-$\ell_2$)\end{tabular} &
      \begin{tabular}[c]{@{}c@{}}CD-Avg\\      (CD-$\ell_1$/CD-$\ell_2$)\end{tabular} &
      \begin{tabular}[c]{@{}c@{}}F-score@$1\%$\\ Avg\end{tabular} \\
    \midrule
    TopNet \cite{tchapmi2019topnet}         & 26.775 / 2.499 & 28.312 / 2.928 & 33.121 / 4.407 & 29.403 / 3.278 & 0.103  \\
    PCN \cite{yuan2018pcn}                  & 27.593 / 2.983 & 28.988 / 3.442 & 34.598 / 5.558 & 30.393 / 3.994 & 0.128  \\
    FoldingNet \cite{yang2018foldingnet}    & 28.356 / 2.887 & 29.832 / 3.290 & 35.356 / 4.968 & 31.181 / 3.715 & 0.088  \\
    GRNet \cite{xie2020grnet}               & 21.247 / 1.554 & 23.757 / 2.287 & 29.426 / 4.169 & 24.810 / 2.670 & 0.208  \\
    ECG \cite{pan2020ecg}                   & 15.283 / 1.255 & 17.596 / 1.759 & 23.534 / 3.267 & 18.804 / 2.094 & \textbf{0.460} \\
    CRN \cite{wang2020cascaded}             & 24.247 / 2.237 & 26.076 / 2.840 & 31.771 / 4.833 & 27.365 / 3.303 & 0.177  \\
    ASFM-Net \cite{xia2021asfm}             & 21.591 / 1.995 & 23.007 / 2.342 & 27.629 / 3.660 & 24.076 / 2.666 & 0.216\\
    PoinTr \cite{yu2021pointr}              & 13.289 / 0.838 & 15.521 / 1.376 & 21.881 / 3.070 & 16.897 / 1.761 & 0.421 \\
    SnowflakeNet \cite{xiang2021snowflakenet} & 15.639 / 1.055 & 18.157 / 1.558 & 24.338 / 3.151 & 19.378 / 1.921 & 0.324 \\
    ProxyFormer \cite{li2023proxyformer}      & 26.954 / 3.065 & 28.313 / 3.516 & 33.716 / 5.530 & 29.660 / 4.037 & 0.139 \\
    AnchorFormer \cite{chen2023anchorformer}  & 16.017 / 1.097 & 17.543 / 1.582  & 21.811 / 2.887 & 18.457 / 1.855 & 0.289 \\
    \textcolor{black}{CRA-PCN~\cite{rong2024cra}}  & 14.431 / 0.835  & 16.245 / 1.448  & 20.117 / 2.801  & 16.931 /1.695  & 0.398 \\
    \textcolor{black}{MMDR~\cite{fei2025multi}}  & \textbf{12.377} /  \textbf{0.806}  &  \textbf{14.598} / \textbf{1.310}  & 19.975 / 2.758  &  \textbf{15.650} /  \textbf{1.625}   &0.334     \\
    \midrule
    \textbf{3DMambaComplete} &14.560 / 0.859 & 21.173 / 2.990 & \textbf{16.225} / \textbf{1.343} & 17.319 / 1.731 & 0.281 \\
    \bottomrule
    \end{tabular}%
    }
  \label{tab:shapnetunseen21_tomm}%
\end{table}%

\subsubsection{Results on ShapeNet34/Unseen21 Dataset}

\begin{figure*}[ht]
    \centering
    \includegraphics[width=\linewidth]{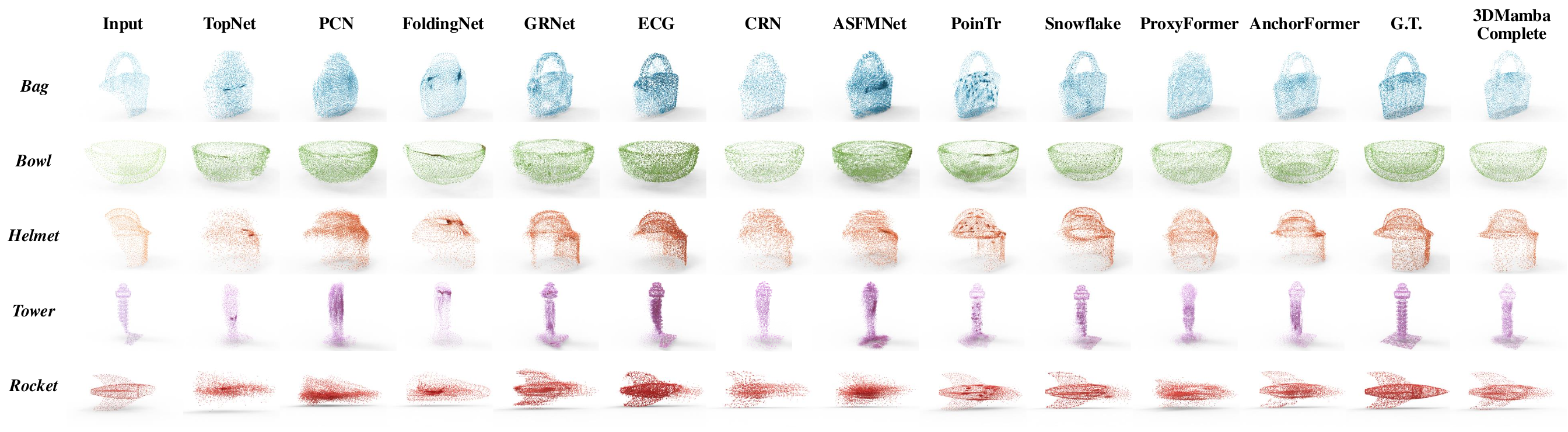}
    \caption{Comparison of different point cloud completion visualization on the ShapeNetUnseen21 \cite{yu2021pointr} dataset.}
    \label{Figure:SN21_comparison}
\end{figure*}

To evaluate the generalization ability of 3DMambaComplete, we conducted experiments on the ShapeNet34 and Unseen21 datasets. As shown in Tables \ref{tab:shapnet34_tomm} and \ref{tab:shapnetunseen21_tomm}, when processing highly incomplete point clouds (under the \emph{H} mask), 3DMambaComplete achieves the best performance on both CD-$\ell_1$ and CD-$\ell_2$ metrics across both seen and unseen categories.
On ShapeNet34, under the \emph{S} and \emph{M} masking settings, the quantitative results of 3DMambaComplete are slightly lower than those of PoinTr. This discrepancy may be attributed to the selective information processing mechanism adopted in our method, which is prone to overfitting in certain scenarios and thus affects the evaluation metrics.
Similarly, on the ShapeNetUnseen21 dataset, under the \emph{S} and \emph{M} masks, the numerical results of 3DMambaComplete are also slightly inferior to those of MMDR.
For the metric F-score@1$\%$, ECG achieve the best performance on both seen and unseen datasets, though its visualisation quality was inferior, as illustrated in Figure~\ref{Figure:SN21_comparison}.
Overall, 3DMambaComplete demonstrates strong generalization capability when completing heavily incomplete point clouds from 21 unseen categories, producing complete and visually coherent results.

\subsection{Ablation Study}

\subsubsection{Ablation Study on Number of the Mamba Blocks.}
\begin{figure}[t]
    \centering
    {\includegraphics[width=\linewidth]
    {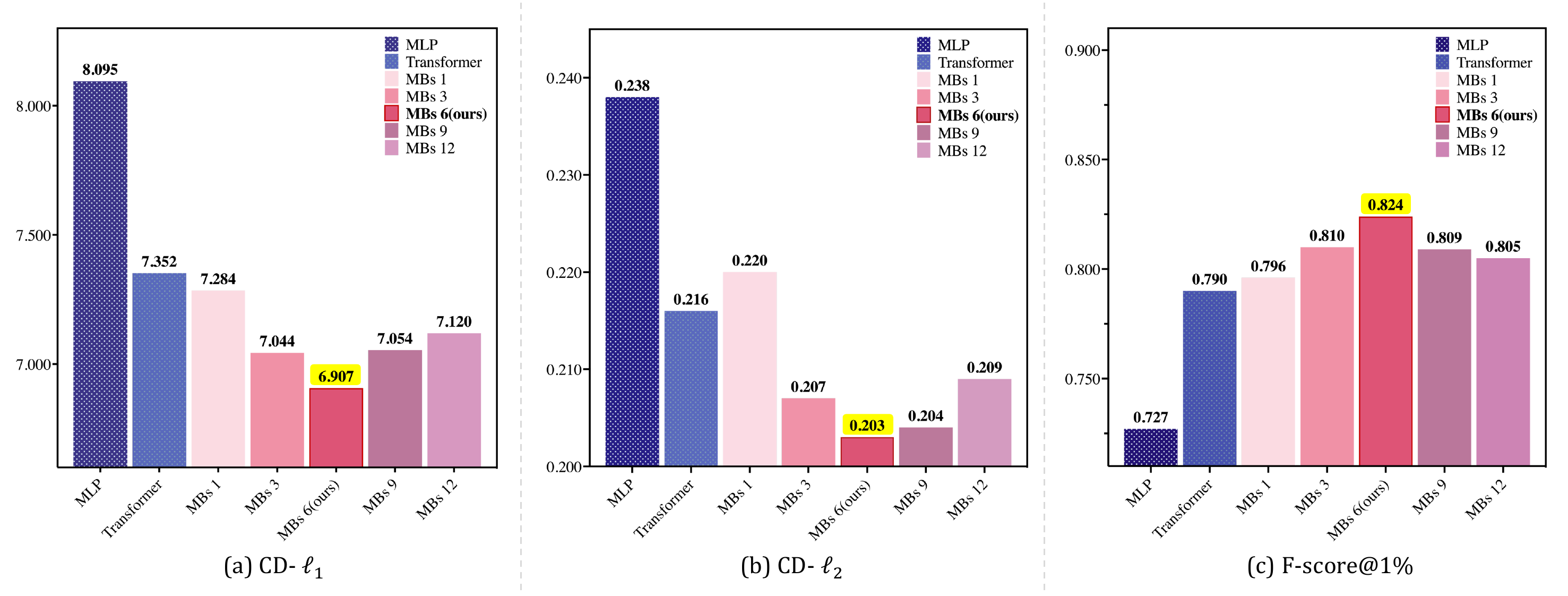}
    \caption{Ablation Experiments on Mamba Blocks in 3DMambaComplete on the PCN Dataset~\cite{yuan2018pcn}. (MBs X denotes X-layer Mamba blocks; Lower CD-$\ell_1$/CD-$\ell_2$ and higher F-score@$1\%$ indicate better performance)}
    \label{fig:ablation_mamba_layers}
    }
\end{figure}

Firstly, we evaluate the impact of Mamba blocks on the model, and the results are detailed in Figure~\ref{fig:ablation_mamba_layers}.
We compare the effects of encoding the point cloud using MLP and Transformer, followed by an evaluation of different numbers of Mamba blocks.
The results show that the collaborative effect of Mamba blocks outperforms the individual effects of MLP and Transformer. 
Furthermore, among the varying numbers of Mamba blocks tested, 3DMambaComplete achieves optimal performance with a total of 6 blocks.
This improvement can be attributed to the insufficient extraction of local point cloud information with fewer Mamba blocks. While increasing the number of Mamba blocks raises computational costs, it also hinders the efficient capture of key information within the Mamba structure.

\subsubsection{Ablation Study on the Different Components.}

\begin{figure}[htbp]
    \centering
    {\includegraphics[width=\linewidth]
    {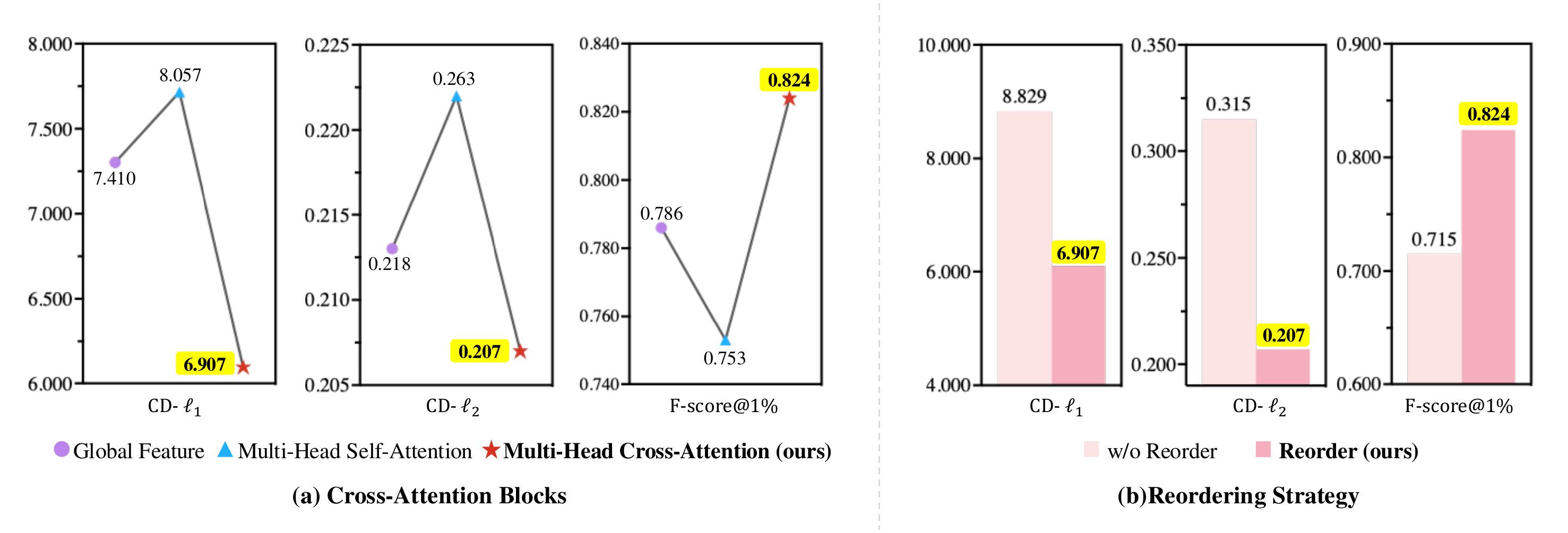}
    \caption{Ablation Experiments on Cross-Attention Blocks and Reordering Strategy in 3DMambaComplete on the PCN Dataset~\cite{yuan2018pcn}. (Lower CD-$\ell_1$/CD-$\ell_2$ and higher F-score@$1\%$ indicate better performance)}
    \label{fig:ablation_components}
    }
\end{figure}

Then, we systematically verify the contribution of each module to the point cloud completion performance. 
The results of the cross-attention module (see Figure \ref{fig:ablation_components} (a)) demonstrate that the introduction of a multi-head cross-attention mechanism at the encoding stage significantly improves the performance of the model on all three evaluation metrics, surpassing self-attention and global feature-based approaches. 
This confirms that cross-attention can reconstruct missing regions more accurately through cross-sequence feature interactions, and particularly excels in handling complex geometric structures. 
In contrast, self-attention lacks cross-sequence feature fusion, limiting its ability to model correlations between local details and global structures. 
As a result, cross-attention is more compatible with point cloud completion and provides an efficient structure-aware solution for high-quality reconstruction.

Following this, ablation experiments on the reordering strategy (see Figure \ref{fig:ablation_components} (b)) show that the model employing the reordering strategy decisively beats the model not employing the strategy on all three metrics. This suggests that by converting the 3D point cloud data into an ordered sequence more suitable for the Mamba framework, the reordering strategy enhances the model's ability to extract features from the point cloud, thereby optimizing the quality of the finished results.

Lastly, we conduct ablation studies with the deformation strategy and loss function (see Table~\ref{tab:ablation_different_components_tomm}). We found that not using the extended loss function (I) or only using simple folding (II) performs worse than the full 3DMambaComplete (III). It indicates that the deformation module using the canonical 2D mesh can reconstruct the point cloud more efficiently, while the newly constructed extended loss function further improves the fine-grained features of the reconstructed point cloud. Therefore, the synergistic effect of these modules greatly increases the overall performance of the model and provides reliable technical support for high-quality point cloud completion.

\begin{table}[htbp]
  \centering
  \caption{Ablation Experiments on the Spread Module Components in 3DMambaComplete on the PCN Dataset~\cite{yuan2018pcn}.}
    \resizebox{0.85\linewidth}{!}{%
    \begin{tabular}{c|ccc|ccc}
    \toprule
          & \multicolumn{1}{c}{Deforming Module} & \multicolumn{1}{c}{Folding Module} & \multicolumn{1}{c|}{Loss Function} & \multicolumn{1}{c}{CD-$\ell_1$} & \multicolumn{1}{c}{CD-$\ell_2$} & \multicolumn{1}{c}{F-score@$1\%$} \\
    \midrule
    I                   & \CheckmarkBold     &  -     &   -    & 7.302 & 0.213 & 0.798 \\
    II                  &  -                & \CheckmarkBold      &  \CheckmarkBold      & 7.716 & 0.222 & 0.763 \\
    \textbf{III (ours)} & \CheckmarkBold       & -      & \CheckmarkBold      & \textbf{6.907} & \textbf{0.207} & \textbf{0.824} \\
    \bottomrule
    \end{tabular}%
    }
  \label{tab:ablation_different_components_tomm}%
\end{table}%

\subsubsection{Taming 3DMambaComplete for Long-sequence Point Clouds.}
\begin{figure}[t]
    \centering
    {\includegraphics[width=0.65\linewidth]
    {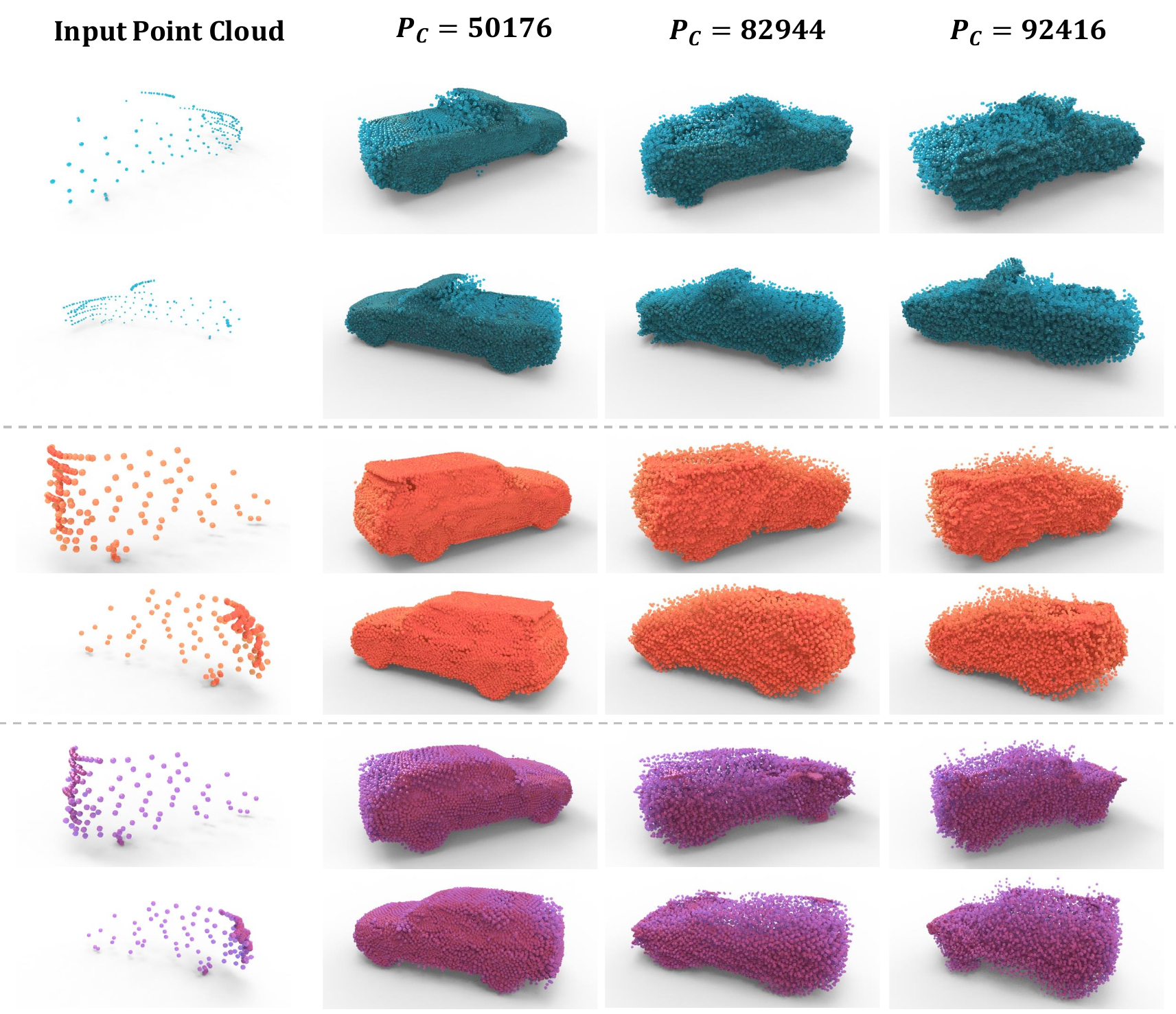}
    \caption{Comparison of different long sequence point cloud complements based on KITTI datasets~\cite{geiger2013vision}. $P_C$ is the number of reconstructed point clouds.}
    \label{fig:long_sequence_vis}
    }
\end{figure}

To demonstrate the effectiveness of 3DMambaComplete towards long-sequence point clouds, we select several point clouds from the KITTI dataset~\cite{geiger2013vision} for reconstruction visualization. 
We try to reconstruct point clouds with up to 92,416 points, and the corresponding results are shown in Figure~\ref{fig:long_sequence_vis}.
Even though the partial point clouds consist of only 2048 points, the predicted complete point clouds exhibit similar reconstruction quality at different numbers of points. 
This robust consistency underscores the exceptional processing capabilities of 3DMambaComplete when handling long sequences of point clouds.

\subsubsection{Ablation Study on the Impact of Complexity Analysis.}
\begin{figure}[t]
    \centering
    {\includegraphics[width=\linewidth]
    {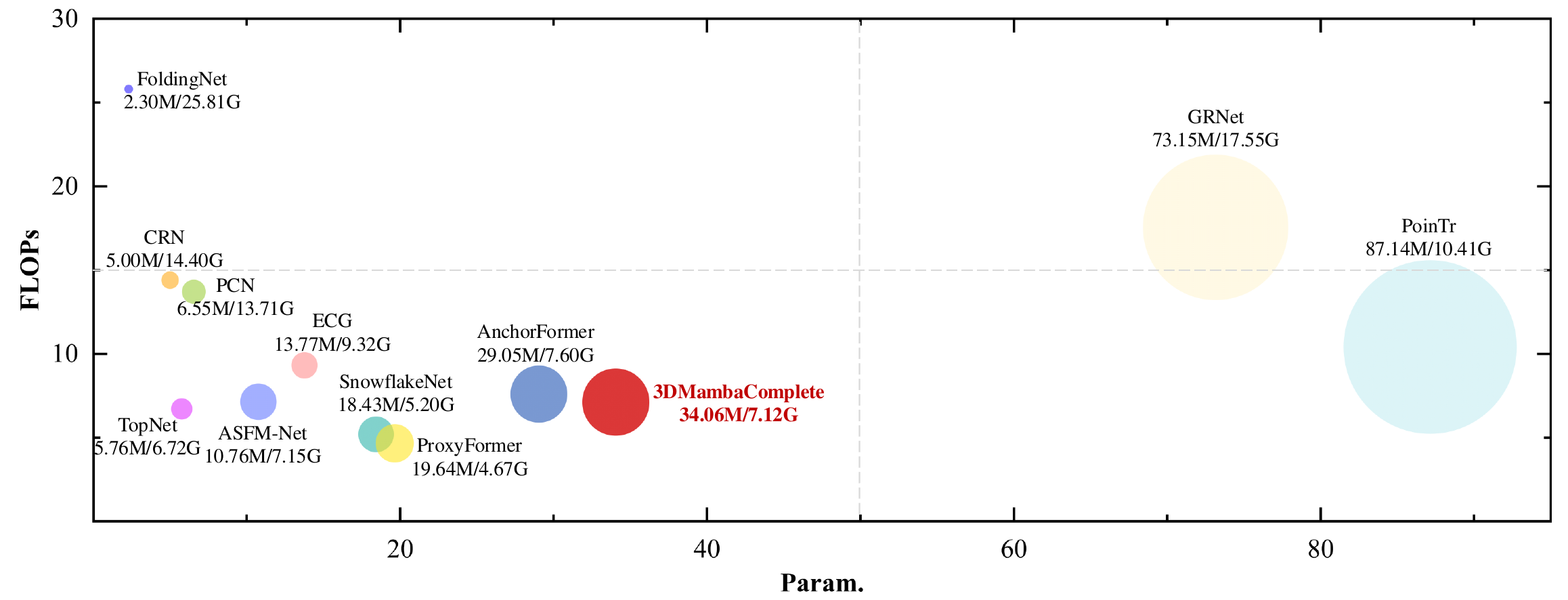}
    \caption{Comparative Analysis of Parameters and FLOPs for Different Methods on the PCN Dataset~\cite{yuan2018pcn}.}
    \label{fig:ablation_flops}
    }
\end{figure}

Finally, we perform a comprehensive evaluation of the number of parameters versus FLOPs for the different methods and summarize the results in Figure~\ref{fig:ablation_flops}.
Regarding the number of parameters, 3DMambaComplete ranks second, with fewer parameters than PoinTr~\cite{yu2021pointr} and GRNet~\cite{yang2018foldingnet}.
The model parameters are largely attributed to the inclusion of innovative components such as HyperPoints and Mamba Blocks. 
Despite this, 3DMambaComplete demonstrated significant efficiency in FLOPs utilization.
Considering the excellent performance and enhanced FLOPs, the model parameter count of 3DMambaComplete are affordable.

\section{Discussion}

\subsection{Limitation}
\begin{figure}[t]
    \centering
    {\includegraphics[width=0.9\linewidth]
    {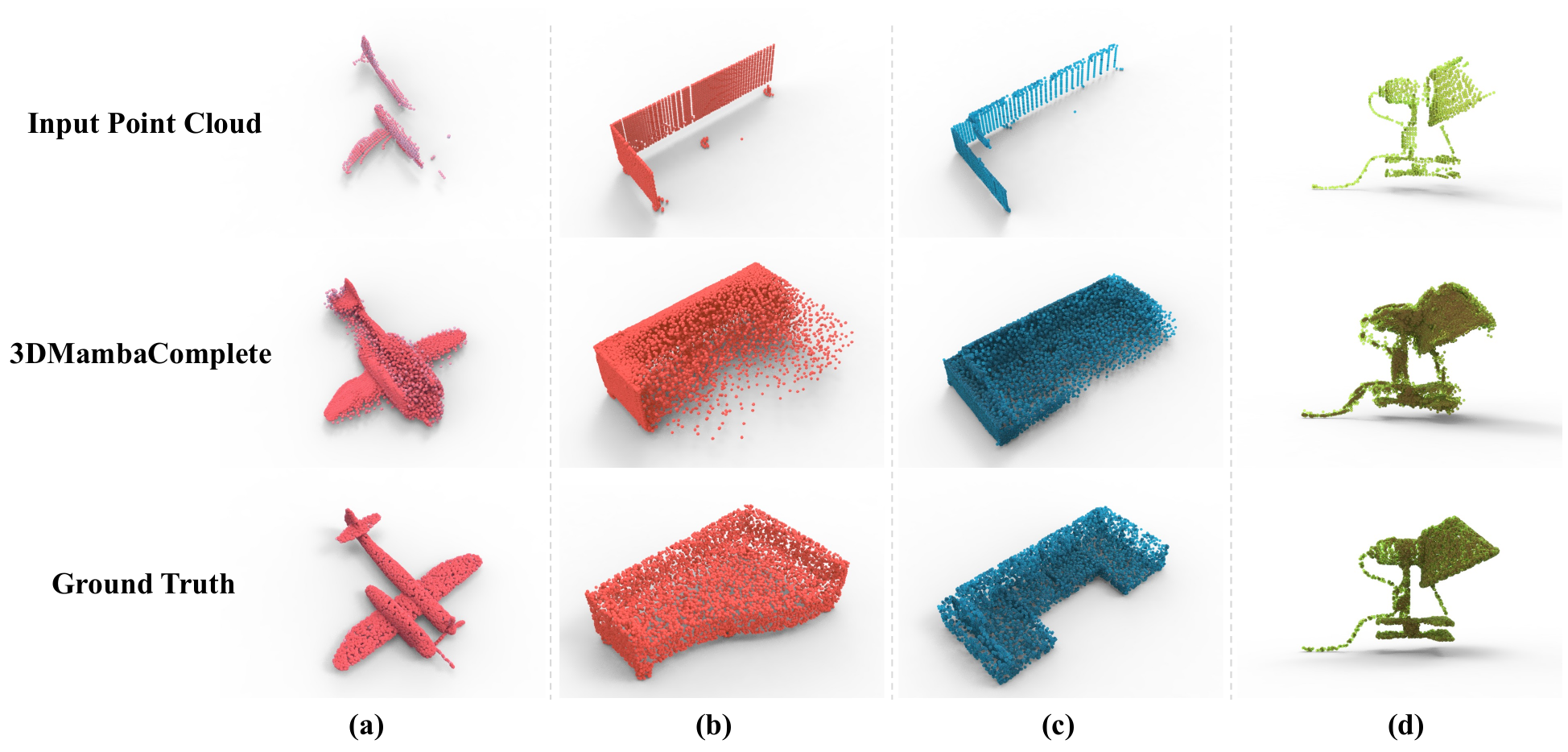}
    \caption{Failure cases of 3DMambaComplete on PCN dataset~\cite{yuan2018pcn}. (a) Asymmetry distortion on airplane wings; (b-c) Density sensitivity causing geometric loss; (d) Topological fracture in lamp mesh reconstruction.}
    \label{fig:failure_case}
    }
\end{figure}

Although 3DMambaComplete achieves excellent performance in point cloud completion with high-fidelity results and reduced computational cost, it still faces several limitations in specific complex scenarios. 
We identify three typical failure cases:
First, asymmetric geometric distortion. When processing asymmetric or irregular objects, such as the airplane in Figure~\ref{fig:failure_case} (a), the model relies too heavily on geometric priors from the training data. The SSM mechanism in Mamba filters out atypical local features, often forcing asymmetric objects into symmetrical shapes and reducing reconstruction accuracy.
Second, density sensitivity. As shown in Figure~\ref{fig:failure_case} (b)(c), 3DMambaComplete produces unsatisfactory results when the input point cloud has missing details or uneven density. The FPS sampling in the HyperPoint Generation Module loses critical geometric information in sparse regions, causing over-concentration of points in dense areas. Meanwhile, the selective SSM emphasizes high-frequency local features over global structure, further shifting predictions toward higher-density zones.
Third, failure in reconstructing complex topologies. For objects with intricate structures, such as the perforated lamp in Figure~\ref{fig:failure_case} (d), the method often yields poor results. The MLP-based Point Deformation Module fails to capture fine topological connectivity, leading to incorrect hole completion or breaks in thin structures.

\subsection{Future Work}
Currently, 3DMambaComplete uses the Mamba framework and intermediate representations such as HyperPoint, which have demonstrated strong modeling capabilities for incomplete point clouds to some extent. However, its capabilities are still limited by the distribution of input data and the structural assumptions of the network itself. Based on the above failure cases, we may advance the development of point cloud completion technology in the following ways.

\begin{itemize}
	\item Uncertainty-Aware Generation: Uncertainty-Aware Generation: Introduce probabilistic frameworks such as diffusion models or variational autoencoders to model the ambiguity and diversity in completions, thereby reducing mode collapse in ill-posed solution spaces.
	\item Geometry-Aware Adaptive Sampling: Develop sampling methods that adapt to local geometry to enhance the robustness and detail preservation in non-uniform point clouds, alleviating issues caused by uniform sampling strategies like FPS.
	\item Explicit Topology Modeling: Integrate structure-aware representations such as graph neural networks or neural implicit surfaces (e.g., SDF) to enforce topological continuity and geometric consistency, improving the reconstruction of complex boundaries and fine details.
\end{itemize}

\section{Conclusion}
This study proposes 3DMambaComplete, a point cloud completion framework based on the State Space Model (SSM) Selective architecture, i.e., Mamba, which is applied for the first time to point cloud completion. 
Leveraging the linear complexity of Mamba, 3DMambaComplete significantly reduces computational costs while efficiently extracting global features, addressing the limitations of traditional Transformer-based methods, such as high computational complexity and detail loss in pooling operations. 
The method introduces discriminative nodes HyperPoints and dynamic offsets to achieve high-quality point cloud reconstruction. 
Specifically, the HyperPoint Generation Module encodes key information using the Mamba Encoder, the HyperPoint Spread Module disperses features with dynamic offsets to avoid aggregation, and the Point Deformation Module transforms a 2D mesh into a detailed 3D structure. 
Experiments show that 3DMambaComplete outperforms existing methods in both quantitative and qualitative evaluations and excels in handling long-sequence point clouds. 
Ablation studies validate the effectiveness of Mamba blocks, the reordering strategy, and the deformation strategy.
In conclusion, by combining the efficiency of Mamba with innovative components, 3DMambaComplete significantly improves the quality and efficiency of point cloud completion, providing an effective solution for handling large-scale and complex point cloud data in the future.

\bibliographystyle{ACM-Reference-Format}
\bibliography{tomm_ref}




\end{document}